\documentclass[
twocolumn,
]{ceurart}

\usepackage{latexsym}
\usepackage[utf8]{inputenc}
\usepackage{microtype}
\usepackage{inconsolata}
\usepackage{graphicx}
\usepackage{changepage}
\usepackage{hyperref}
\usepackage{caption}
\usepackage{subcaption}
\usepackage{multirow}
\usepackage{array}
\usepackage{tikz}
\usepackage{enumitem}
\usepackage[T1]{fontenc}
\usepackage{type1cm}
\usepackage{type1ec}
\usepackage{xcolor}
\usepackage{boldline} 

\usepackage{tcolorbox}
\usepackage{verbatim}

\DeclareMathOperator*{\argmax}{argmax}

\usepackage{eucal}

\renewcommand{\mathcal}{\EuScript}

\let\xx\textbf

\newcommand{\quotes}[1]{``#1''}

\usepackage{bm}

\newcommand{\mathbcal}[1]{\bm{\mathcal{#1}}}

\definecolor{mybluex}{HTML}{0165FC}
\definecolor{myblueb}{HTML}{0A3161}
\definecolor{myredb}{HTML}{B31942}

\usepackage{listings}
\begin{document}

 
\copyrightyear{2024}
\copyrightclause{Copyright for this paper by its authors.
  Use permitted under Creative Commons License Attribution 4.0
  International (CC BY 4.0).}

\conference{RecSys in HR'24: The 4th Workshop on Recommender Systems for Human Resources, in conjunction with the 18th ACM Conference on Recommender Systems, October 14--18, 2024, Bari, Italy.}

\title{MELO: An Evaluation Benchmark for Multilingual Entity Linking of Occupations}

\author[]{Federico Retyk}[
    email=machinelearning@avature.net
]
\author[]{Luis Gascó}[]
\author[]{Casimiro Pio Carrino}[]
\author[]{Daniel Deniz}[]
\author[]{Rabih Zbib}[]
\address[]{Avature Machine Learning}

\begin{abstract}
We present the Multilingual Entity Linking of Occupations (MELO) Benchmark, a new collection of 48 datasets for evaluating the linking of entity mentions in 21 languages to the ESCO Occupations multilingual taxonomy. MELO was built using high-quality, pre-existent human annotations. We conduct experiments with simple lexical models and general-purpose sentence encoders, evaluated as bi-encoders in a zero-shot setup, to establish baselines for future research. The datasets and source code for standardized evaluation are publicly available at \url{https://github.com/Avature/melo-benchmark}.

\end{abstract}


\maketitle


\section{Introduction}
\label{sec:introduction}

The current trend in the digital transformation of human resources~(HR) processes is the integration of artificial intelligence~(AI) components that can improve automation and operational efficiency. These systems often need to process input data in the form of natural language text, which can be noisy and diverse in terms of language and other domain-specific aspects.

One common approach to deal with this challenge is the application of entity linking (EL) methods. EL helps normalizing input data into standardized entities within well-curated taxonomies. These taxonomies facilitate interoperability across different systems and, when multilingual, enable the integration of information across languages. In highly specialized domains like HR and recruiting, the development of EL methods faces significant challenges, particularly when training resources are scarce or nonexistent~\cite{logeswaran-etal-2019-zero, wu-etal-2020-scalable}. These challenges are further amplified in multilingual environments~\cite{botha-etal-2020-entity, fu-etal-2020-design}. Therefore, achieving accurate entity resolution across languages is key to ensuring the consistency and effectiveness of digitalized HR systems in a global setting.

Previous research in the application of AI within the HR domain has made extensive use of taxonomies, such as occupation and skill classifications~\cite{groot2021job, tu2022humanintheloop, avlonitis2023careerpath, decorte2023careerpath, zhang-etal-2022-skillspan, zhang-etal-2023-escoxlm}. These HR-specific taxonomies have been used for normalizing raw data~\cite{zhang-etal-2024-entity, senger-etal-2024-deep, zhang-etal-2022-kompetencer, decorte2023negsampl, zhang2022skillextr, lake2022jobclass, decorte2023skillsgpt}, removing noise and enabling AI models to operate on standardized information, which in turn leads to more accurate and reliable outcomes. Substantial progress has been made, particularly in the normalization of occupational data~\cite{giabelli2021weta, giabelli2024phdthesis, decorte2021, yamashita2023james, vrolijk2023enhancing}. However, despite these advancements, there is still a surprising lack of high-quality public evaluation benchmarks for measuring progress consistently in this important area.

To address this gap, we propose the Multilingual Entity Linking of Occupations (MELO) Benchmark, a new collection of 48 datasets designed to evaluate multilingual EL tasks. This benchmark leverages pre-existing, high-quality human annotations and covers 21 languages. Furthermore, we present an experimental study using the new benchmark to evaluate the performance of both simple lexical baselines and existing deep learning models employed as zero-shot bi-encoders. Our goal is for MELO to serve as a valuable resource for advancing research and fostering innovation in this field.

The main contributions of this work are:

\begin{itemize}

  \item We introduce the MELO Benchmark, a suite of 48 datasets involving monolingual, cross-lingual, and multilingual tasks in 21 languages. Each dataset corresponds to an entity linking task framed as a ranking problem, where queries and corpus elements are occupation names taken from a source and a target taxonomy, respectively, and binary-relevance annotations are derived from high-quality crosswalks between the taxonomies. Additionally, we release code for standardizing the evaluation of models on this benchmark.
  
  \item We provide experimental results for both simple lexical systems and state-of-the-art deep learning models evaluated as zero-shot bi-encoders on MELO, to serve as baselines for future research. We find that, while the lexical baselines perform fairly well, the semantic baselines generally achieve better results, particularly in cross-lingual tasks. However, there remains significant room for improvement.

\end{itemize}

To the best of our knowledge, MELO is the first public evaluation benchmark to address the task of multilingual entity linking in the HR domain.


\section{Background}
\label{sec:background}

In this Section, we introduce the context necessary for understanding the subsequent task definitions (\S\ref{sec:task}), the methodology employed in constructing the benchmark (\S\ref{sec:datasets}), and the related work (\S\ref{sec:related-work}).


\noindent \textbf{Entity Linking}.~~~Given a knowledge base $\mathbcal{E}$ and a query mention $q$, the task of Entity Linking (EL) involves identifying the correct entity $e \in \mathbcal{E}$ to which the mention is referring. In principle, the structure of the knowledge base $\mathbcal{E}$ can range from a flat catalog of unrelated entities to a complex and heterogeneous ontology. In this work we focus on taxonomies of a single type of entity (i.e.~occupations).

Inspired by the multilingual formulation proposed by Botha~et~al.~\cite{botha-etal-2020-entity}, we consider each entity $e$ as a language-agnostic concept with associated language-specific textual information. For each language $l$ in a set of supported languages $\mathbcal{L}^{tax}$, any entity may have a set of names (synonymous between each other), a description, and example sentences where the concept is used. The query $q$ is a text string in some language $l^{q}$, with no prior assumptions about the relationship between $l^{q}$ and the set $\mathbcal{L}^{tax}$ of supported languages\footnote{For example, setting $\mathbcal{L}^{tax}=\{ l^{q} \}$ would result in a monolingual task, and $\mathbcal{L}^{tax}=\{ l^{x} \}$ with $l^{q} \neq l^{x}$ involves a cross-lingual task. More generally, a set $\mathbcal{L}^{tax}$ with higher cardinality can define a multilingual task.}.

In principle, the system may receive a query mention $q$ that refers to an entity that does not exist in the taxonomy, or it may not refer to any entity at all. This problem, known as out-of-KB or NIL prediction~\cite{zhu-etal-2023-learn}, falls outside the scope of this work. Additionally, it is typical in the EL community to allow the system to know the textual context in which the mention occurs, aiding in the resolution of ambiguity~\cite{gupta-etal-2017-entity}. This aspect is also beyond the scope of our work, as the data we use to build our datasets only includes unnormalized occupation names as queries.

Entity linking can be framed as a ranking task~\cite{zheng-etal-2010-learning}: given a query $q$, the system produces a score $s(q, e)$ for each $e \in \mathbcal{E}$ and the predicted entity $\hat{e}$ is computed as:

\begin{equation*}
    \hat{e}(q) = \argmax_{e \; \in \; \mathcal{E}} s(q, e)
    \label{eq:argmax-score}
\end{equation*}

\noindent and rank-based evaluation metrics can be used to study the performance. A typical approach to this task breaks it into two stages. The first is the \emph{Candidate Generation Stage}, where an initial ranking is obtained using a low-latency method, trying to optimize for recall. In the second stage, the \emph{Re-ranking Stage}, a more costly but higher-precision method is applied to evaluate the top elements in the preliminary rank.

Obtaining annotated data for training such systems is costly, particularly for tasks involving custom taxonomies or low-resource languages~\cite{fu-etal-2020-design}. To mitigate this problem, many techniques have been proposed for leveraging transfer learning to obtain good performance in zero-shot EL scenarios~\cite{logeswaran-etal-2019-zero, wu-etal-2020-scalable}. State-of-the-art methods typically use a bi-encoder for the candidate generation stage, and a cross-encoder for the re-ranking stage.



\noindent \textbf{Multilingual Taxonomies}.~~~For the purposes of this work, we define a taxonomy $\mathbcal{E}$ as a directed acyclic graph (DAG) where nodes are concepts and edges represent binary IS-A relationships~\cite{brachman1983taxonomies} between concepts. The tail concept (child) is a hyponym of the head concept (parent) and therefore represents a narrower meaning. Conversely, the parent is a hypernym of the child and represents a broader meaning, i.e.\ a category to which the child belongs. Concepts are allowed to have many parents.

In a multilingual taxonomy, concepts are language-agnostic but they have language-specific properties, such as a set of names, a description, or usage examples. In other words, every concept has one set of names for each language supported in the taxonomy. The set of names for a concept for a language are considered synonyms between each other. If a lexical entry is attached to more than one concept, this implies polysemy.


\noindent \textbf{Occupation Taxonomies}.~~~Several public occupation taxonomies were developed to classify, standardize, and organize information related to job titles and roles found in the workforce.

One popular and influential occupations taxonomy is the European Skills, Competences, Qualifications, and Occupations (ESCO) ontology, a collection of multilingual and interrelated taxonomies created and maintained by the European Union~\cite{levrang2014, escoHandbook}. It includes $3{,}039$ occupation concepts in its latest version, each with names and definitions (descriptions) in $28$ languages. Every concept has one or more names in every supported language. The names are compliant with the terminological guidelines defined by ESCO~\cite{escoGuidelines}. All the names of a particular concept in a particular language are considered synonyms with each other. Also, for a particular concept, the language-specific name sets can be considered parallel data from a translation point of view.

Another important example is the O*NET-SOC taxonomy. The Occupational Information Network (O*NET) is developed and maintained by the United States government~\cite{onet2006, onet2016} to standardize information relevant to the labor market, based on the 2018 Standard Occupational Classification (SOC) system\footnote{https://www.bls.gov/soc/}. It contains information in English about $1{,}016$ occupations, each with a set of names and a description.

Additionally, many other countries have developed their own national taxonomies or terminologies for occupations. For example, the Federal Employment Agency in Germany developed the \textit{Klassifikation der Berufe} 2010 (KldB 2010) which is a terminology used to standardize the information in the German language about occupations~\cite{paulus2013klassifikation}.

To achieve interoperability between some of these taxonomies, mappings ---also called \textbf{crosswalks}--- were developed and made public. These mappings establish an alignment between two given taxonomies. In particular, the European Union published many crosswalks~\cite{escoOnetCrosswalk} that map concepts from national taxonomies, which are typically monolingual, into ESCO. The process described in Section~\ref{sec:datasets} uses this information as a gold standard to create the datasets for the MELO Benchmark.


\begin{table*}[htbp]
\footnotesize
\begin{center}
{\caption{
Datasets in the MELO Benchmark.\ \textsuperscript{\color{mybluex} \textdagger} USA-en-xx is the only multilingual dataset. $\color{mybluex} \mathbcal{L}^{xx}$ denotes the set of languages of the elements in the corpus: English, German, Spanish, French, Italian, Dutch, Portuguese, and Polish.
}\label{tab:corpora-main}}
\begin{tabular}{llcrccr}

  \hlineB{3}
  \multicolumn{1}{c}{\multirow{2}{*}{Task Name}} &
  \multicolumn{1}{c}{\multirow{2}{*}{Source Taxonomy}} &
  \multicolumn{2}{c}{Queries} &
  \multicolumn{1}{c}{\multirow{2}{*}{Target Taxonomy}} &
  \multicolumn{2}{c}{Corpus Elements}
  \\
  \cline{3-4} \cline{6-7} 
  \multicolumn{1}{c}{} &
  \multicolumn{1}{c}{} &
  \multicolumn{1}{c}{Language} &
  \multicolumn{1}{c}{\#} &
  \multicolumn{1}{c}{} &
  \multicolumn{1}{c}{Language} &
  \multicolumn{1}{c}{\#}
  \\
  \hlineB{3}
  USA-en-en  & O*NET            & en & 633    & ESCO v1.1.0 & en       & 33,813    \\
  USA-en-xx\textsuperscript{\color{mybluex} \textdagger}  & O*NET            & en & 633    & ESCO v1.1.0  & $\color{mybluex} \mathbcal{L}^{xx}$ & 150,140   \\ 
  \hline
  AUT-de-de  & Austria          & de & 1,120  & ESCO v1.1.0  & de       & 19,782    \\
  AUT-de-en  & Austria          & de & 1,120  & ESCO v1.1.0  & en       & 33,813    \\
  BEL-fr-fr  & Belgium          & fr & 328    & ESCO v1.0.3  & fr       & 15,227    \\
  BEL-fr-en  & Belgium          & fr & 328    & ESCO v1.0.3  & en       & 33,609    \\
  BEL-nl-nl  & Belgium          & nl & 328    & ESCO v1.0.3  & nl       & 24,070    \\
  BEL-nl-en  & Belgium          & nl & 328    & ESCO v1.0.3  & en       & 33,609    \\
  BGR-bg-bg  & Bulgaria         & bg & 4,438  & ESCO v1.0.3  & bg       & 21,082    \\
  BGR-bg-en  & Bulgaria         & bg & 4,438  & ESCO v1.0.3  & en       & 33,609    \\
  CZE-cs-cs  & Czechia          & cs & 988    & ESCO v1.0.9  & cs       & 13,333    \\
  CZE-cs-en  & Czechia          & cs & 988    & ESCO v1.0.9  & en       & 33,583    \\
  DEU-de-de  & Germany          & de & 1,779  & ESCO v1.0.3  & de       & 19,135    \\
  DEU-de-en  & Germany          & de & 1,779  & ESCO v1.0.3  & en       & 33,609    \\
  DNK-da-da  & Denmark          & da & 734    & ESCO v1.0.8  & da       & 10,410    \\
  DNK-da-en  & Denmark          & da & 734    & ESCO v1.0.8  & en       & 33,583    \\
  ESP-es-es  & Spain            & es & 1,580  & ESCO v1.0.8  & es       & 16,502    \\
  ESP-es-en  & Spain            & es & 1,580  & ESCO v1.0.8  & en       & 33,583    \\
  EST-et-et  & Estonia          & et & 1,068  & ESCO v1.0.8  & et       & 4,956     \\
  EST-et-en  & Estonia          & et & 1,068  & ESCO v1.0.8  & en       & 33,583    \\
  FRA-fr-fr  & France           & fr & 1,435  & ESCO v1.0.9  & fr       & 15,217    \\
  FRA-fr-en  & France           & fr & 1,435  & ESCO v1.0.9  & en       & 33,583    \\
  HRV-hr-hr  & Croatia          & hr & 2,347  & ESCO v1.0.3  & hr       & 17,390    \\
  HRV-hr-en  & Croatia          & hr & 2,347  & ESCO v1.0.3  & en       & 33,609    \\
  HUN-hu-hu  & Hungary          & hu & 362    & ESCO v1.0.8  & hu       & 16,923    \\
  HUN-hu-en  & Hungary          & hu & 362    & ESCO v1.0.8  & en       & 33,583    \\
  ITA-it-it  & Italy            & it & 362    & ESCO v1.0.8  & it       & 16,199    \\
  ITA-it-en  & Italy            & it & 362    & ESCO v1.0.8  & en       & 33,583    \\
  LTU-lt-lt  & Lithuania        & lt & 3,849  & ESCO v1.0.8  & lt       & 17,824    \\
  LTU-lt-en  & Lithuania        & lt & 3,849  & ESCO v1.0.8  & en       & 33,583    \\
  LVA-lv-lv  & Latvia           & lv & 3,251  & ESCO v1.0.8  & lv       & 9,733     \\
  LVA-lv-en  & Latvia           & lv & 3,251  & ESCO v1.0.8  & en       & 33,583    \\
  NLD-nl-nl  & Netherlands      & nl & 2,605  & ESCO v1.0.3  & nl       & 24,070    \\
  NLD-nl-en  & Netherlands      & nl & 2,605  & ESCO v1.0.3  & en       & 33,609    \\
  NOR-no-no  & Norway           & no & 96      & ESCO v1.0.8  & no       & 7,821     \\
  NOR-no-en  & Norway           & no & 96      & ESCO v1.0.8  & en       & 33,583    \\
  POL-pl-pl  & Poland           & pl & 1,937  & ESCO v1.0.3  & pl       & 8,879     \\
  POL-pl-en  & Poland           & pl & 1,937  & ESCO v1.0.3  & en       & 33,609    \\
  PRT-pt-pt  & Portugal         & pt & 379    & ESCO v1.0.3  & pt       & 11,671    \\
  PRT-pt-en  & Portugal         & pt & 379    & ESCO v1.0.3  & en       & 33,609    \\
  ROU-ro-ro  & Romania          & ro & 3,273  & ESCO v1.0.8  & ro       & 14,833    \\
  ROU-ro-en  & Romania          & ro & 3,273  & ESCO v1.0.8  & en       & 33,583    \\
  SVK-sk-sk  & Slovakia         & sk & 2,040  & ESCO v1.0.8  & sk       & 12,899    \\
  SVK-sk-en  & Slovakia         & sk & 2,040  & ESCO v1.0.8  & en       & 33,583    \\
  SVN-sl-sl  & Slovenia         & sl & 3,222  & ESCO v1.0.8  & sl       & 15,487    \\
  SVN-sl-en  & Slovenia         & sl & 3,222  & ESCO v1.0.8  & en       & 33,583    \\
  SWE-sv-sv  & Sweden           & sv & 2,883  & ESCO v1.1.1  & sv       & 7,506     \\
  SWE-sv-en  & Sweden           & sv & 2,883  & ESCO v1.1.1  & en       & 33,802    \\
  \hlineB{3}
\end{tabular}
\end{center}
\end{table*}


\section{Task}
\label{sec:task}

As mentioned already, the task consists of multilingual Entity Linking of occupations into the ESCO taxonomy, which we denote by $\mathbcal{E}$. Given a query mention $q$, which is a text string expressing the non-normalized name of an occupation without surrounding context, we need to find the best semantic match in ESCO, namely the correct entity $e \in \mathbcal{E}$. Every occupation in the taxonomy has textual information in all languages $l \in \mathbcal{L}^{tax}$. The query is expressed in language $l^q$, which we make no prior assumptions about.

For evaluation, we operationalize the task as a ranking problem with binary-relevance annotations, where a query $q$ is used to rank all the strings $c_{i}$ in a corpus $\mathbcal{C}$. The corpus is a collection of lexical terms denoting occupation names, and it is derived from the taxonomy~$\mathbcal{E}$.

To build the corpus $\mathbcal{C}$, we first define the set of target languages for the corpus, as a subset $\mathbcal{L}^{c} \subset \mathbcal{L}^{tax}$. Then, we collect every surface form (name) for every occupation corresponding to those languages. That is, starting from an empty set, we traverse $\mathbcal{E}$ and, for each occupation $e$, we add every name available in any language in $\mathbcal{L}^{c}$. As a result, $\mathbcal{C}$ is the collection of every name of every occupation in every target language.

The annotations consist of the set of relevant corpus elements for each query. Given the correct entity $e$ for a query $q$, then those corpus elements $c_{i}$ that were obtained from the surface forms of $e$ are considered to be relevant, while any other element in the corpus is considered irrelevant.

Because the goal is to find the relevant concept $e$ in the taxonomy for the given query (i.e.~to solve the entity linking formulation of the task), obtaining at least one surface form $c_{i}$ associated with the relevant concept at the top of the ranking is sufficient for correctly performing the task. In other words, when ranking the corpus elements for a query, the position in the ranking of the highest-ranked relevant surface form is the measure we aim to evaluate. For this reason, we evaluate the baseline models with the following metrics: mean reciprocal rank~(MRR) and top-$k$ accuracy~(A@$k$).


\section{Datasets}
\label{sec:datasets}

The MELO Benchmark consists of 48 datasets, where each is an instance of the ranking task as described in Section~\ref{sec:task}. While the set of queries differs among the datasets, the target taxonomy is always ESCO Occupations. Although the underlying concepts in the corpus are the same, the surface forms ---specifically, the occupation names--- vary across datasets, since they are presented in different subsets of ESCO languages.

We leverage existing crosswalks\footnote{\url{https://esco.ec.europa.eu/en/use-esco/eures-countries-mapping-tables}}, which are high-quality mappings between ESCO Occupations and other taxonomies~\cite{escoMappings, escoOnetCrosswalk}, to build the datasets. Two datasets are derived from the mapping between ESCO and the O*NET-SOC Taxonomy, while the remaining ones are derived from the mapping between ESCO and the official occupation terminologies from several European countries. While ESCO is a multilingual taxonomy, the national terminologies are monolingual. Elements between the taxonomies are assigned SKOS relationships~\cite{skos2009} such as \textit{exact match}, \textit{narrow match}, \textit{broad match}, or \textit{close match}.

For each crosswalk, we build two evaluation datasets: a monolingual dataset and a cross-lingual dataset. In both cases, the set of queries are those elements in the national terminologies (or O*NET) that either have only one \textit{exact match} in ESCO or have zero \textit{exact matches} and only one \textit{narrow match}. Therefore, we are filtering out semantically ambiguous queries, e.g.~if they have more than one \textit{exact matches}, or that can't be assigned to a specific concept in ESCO because they are not specific enough, for example if they only have \textit{broad} or \textit{close matches}.

The language of the set of queries, $l^{q}$, depends on the national terminology. Regarding the languages used for the corpus, we select a different subset of the languages in ESCO for each modality. For the monolingual task we set $\mathbcal{L}^{c}=\{ l^{q} \}$, and for the cross-lingual we set $\mathbcal{L}^{c}=\{ \texttt{English} \}$. Exceptionally, since for O*NET the query language is already English, in this case instead of a cross-lingual task we define a multilingual task, where the corpus languages are English, German, Spanish, French, Italian, Dutch, Portuguese, and Polish (We intentionally include English, the query language.)
As mentioned in the previous Section, the annotations consist of relevancy pairs, where the set of corpus elements that correspond to the correct occupation entity for a particular query are marked as relevant, while all other corpus elements are irrelevant.

To illustrate this with an example, given the national terminology of France, we use the corresponding crosswalk to build two datasets: the monolingual dataset, where both the queries and the corpus elements are in French, and a cross-lingual dataset, where the queries are in French but the corpus elements are in English. We name these datasets \texttt{FRA-fr-fr} and \texttt{FRA-fr-en}, respectively. In Table~\ref{tab:corpora-main} we list all the datasets in the benchmark, with information about the languages and number of elements in their query and corpus element sets. For further detail on the construction and composition of these datasets, as well as example queries and relevant corpus elements, please refer to Appendix~\ref{sec:appendix-details-datasets}.

The benchmark is intended to represent realistic use cases, such as linking mentions into a taxonomy, enriching a custom taxonomy with new synonyms for the existing concepts, or aligning two taxonomies. It is also intended to study the cross-lingual and multilingual capabilities of proposed systems. Using extra information for solving this task, such as context for the mentions or descriptions and examples for the taxonomy concepts, is out of the scope of this work but represents an interesting line of future research that can take advantage of the MELO Benchmark.



\begin{figure*}[t]
\begin{center}
\includegraphics[width=0.92\textwidth]{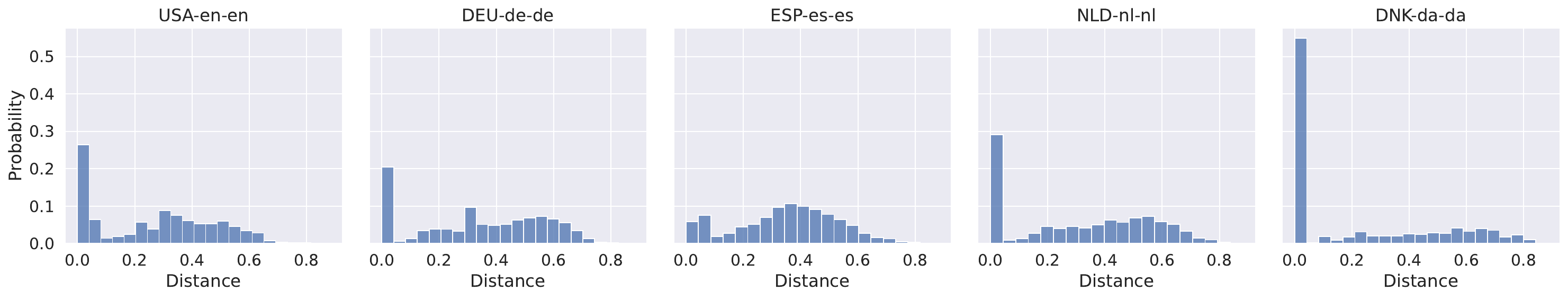}
\caption{Histogram of minimum (normalized) edit distances between each query and the closest relevant corpus element for a selection of monolingual tasks in MELO.}
\label{fig:hist-main-some}
\end{center}
\end{figure*}

To assess the lexical overlap between the surface forms in any national terminology and ESCO, we use the monolingual tasks, and measure the normalized edit distance between each query and the closest relevant corpus element. In Figure~\ref{fig:hist-main-some} we show a histogram with the distribution of such distances in a selection of tasks.

The lexical overlap is considerable in some cases, like with the Danish terminology. In the histogram, a big concentration of examples in the left-most bin implies that many queries are lexically very close to their relevant corpus elements. This, in principle, would make these tasks easier to solve using simple lexical scoring functions. In Appendix~\ref{sec:appendix-details-datasets} we explain the procedure used to compute the lexical distances and we also present the same analysis for every task in the benchmark.


\section{Experiments}
\label{sec:experiments}

To demonstrate the MELO Benchmark in use, we study the performance of several models when evaluated on the tasks we defined above. We explore both simple lexical baselines and advanced deep learning models using a bi-encoder, zero-shot setting.

\noindent \textbf{Lexical Baselines}.~~~We evaluate the following baselines: edit-distance, word-level TF-IDF, word-level TF-IDF on lemmas, char-level TF-IDF, char-level TF-IDF on lemmas, BM25, and BM25 on lemmas. These models rely on surface-level text features.

\noindent \textbf{Semantic Baselines}.~~~Additionally, we provide results for zero-shot evaluations using state-of-the-art deep learning models employed as symmetric bi-encoders. Under this setup, we use a sentence encoder to obtain a fixed-size representation for each surface form, and the score for a query and each corpus element is computed as the cosine similarity of their corresponding representations. This allows the system to capture deeper semantic relationships.

We experiment with the following pre-trained models in a zero-shot setup, without fine-tuning or in-context examples: ESCOXLM-R~\cite{zhang-etal-2023-escoxlm}, mUSE-CNN~\cite{yang-etal-2020-multilingual}, a multilingual variant of MPNet~\cite{song2020mpnet}, BGE-M3~\cite{chen2024bge}, GIST-Embedding~\cite{solatorio2024gistembed}, Multilingual E5~\cite{wang2024multilingual}, E5~\cite{wang2023improving, wang2022text}, and the model \texttt{text-embedding-3-large} from OpenAI\footnote{\url{https://openai.com/index/new-embedding-models-and-api-updates}}. This selection of models represents a spectrum of trade-offs between performance and model complexity. We refer the reader to Appendix~\ref{sec:appendix-details-models} and Table~\ref{tab:models-detail} for further details on the models and the inference procedure.


\begin{table*}[htbp]
\footnotesize
\begin{center}
{\caption{
Mean reciprocal rank (MRR) for each model, evaluated in the monolingual and the cross-lingual versions of a selection of tasks in MELO.\ \textsuperscript{\color{mybluex} \textdagger} USA-en-xx is a multilingual dataset, with corpus elements that also cover the language of the query.
}\label{tab:results-main-mrr}}
\begin{tabular}{lcccccccccc}
  \hlineB{3}
  \multirow{2}{*}{Model} &
  \multicolumn{2}{c}{USA} &
  \multicolumn{2}{c}{DEU} &
  \multicolumn{2}{c}{ESP} &
  \multicolumn{2}{c}{NLD} &
  \multicolumn{2}{c}{DNK}
  \\
  \cline{2-11} 
                       & en-en       & en-xx \textsuperscript{\color{mybluex} \textdagger}       & de-de       & de-en       & es-es       & es-en       & nl-nl       & nl-en       & da-da       & da-en       \\ 
  \hlineB{3}
  Edit Distance        & 0.4858      & 0.4889      & 0.4392      & 0.0832      & 0.3297      & 0.0545      & 0.4275      & 0.0952      & 0.5650      & 0.1596      \\ 
  Word TF-IDF          & 0.3250      & 0.3207      & 0.4763      & 0.0388      & 0.2411      & 0.0127      & 0.4714      & 0.0460      & 0.5187      & 0.0398      \\ 
  Word TF-IDF (lemmas) & \xx{0.6056} & \xx{0.5999} & 0.4666      & 0.0391      & 0.4318      & 0.0307      & 0.4674      & 0.0435      & 0.5179      & 0.0404      \\ 
  Char TF-IDF          & 0.5800      & 0.5764      & 0.5442      & \xx{0.1301} & 0.4376      & 0.1238      & \xx{0.4862} & 0.1281      & \xx{0.5809} & \xx{0.1576} \\ 
  Char TF-IDF (lemmas) & 0.5957      & 0.5913      & \xx{0.5474} & 0.1278      & \xx{0.4697} & \xx{0.1347} & 0.4811      & \xx{0.1321} & 0.5801      & 0.1551      \\ 
  BM25                 & 0.2936      & 0.2814      & 0.3377      & 0.0050      & 0.1916      & 0.0073      & 0.4433      & 0.0338      & 0.4987      & 0.0296      \\ 
  BM25 (lemmas)        & 0.6004      & 0.5978      & 0.4473      & 0.0198      & 0.4367      & 0.0275      & 0.4320      & 0.0393      & 0.5125      & 0.0334      \\ 
  \hline
  ESCOXLM-R            & 0.3450      & 0.3426      & 0.4087      & 0.1002      & 0.2476      & 0.0854      & 0.3184      & 0.0829      & 0.3631      & 0.1095      \\ 
  mUSE-CNN             & 0.5532      & 0.5317      & 0.5606      & 0.3138      & 0.4176      & 0.3217      & 0.4255      & 0.2666      & 0.5026      & 0.1680      \\ 
  Paraph-mMPNet        & 0.5876      & 0.5822      & 0.4691      & 0.0916      & 0.3417      & 0.0899      & 0.3831      & 0.0955      & 0.4602      & 0.1148      \\ 
  BGE-M3               & 0.6226      & 0.6301      & 0.6083      & 0.3344      & 0.4927      & 0.3084      & 0.5045      & 0.3033      & 0.5839      & 0.3037      \\ 
  GIST-Embedding       & 0.6431      & 0.6464      & 0.5363      & 0.1325      & 0.3574      & 0.1534      & 0.4487      & 0.1316      & 0.5608      & 0.1348      \\ 
  mE5                  & 0.6563      & 0.6588      & 0.6122      & 0.3858      & 0.5021      & 0.3480      & 0.5059      & 0.3246      & 0.5983      & 0.3325      \\ 
  E5                   & 0.6735      & 0.6777      & 0.6639      & 0.5073      & \xx{0.5557} & 0.4628      & 0.5650      & 0.4133      & 0.6178      & 0.4053      \\ 
  OpenAI               & \xx{0.6842} & \xx{0.6872} & \xx{0.6778} & \xx{0.5518} & 0.5371      & \xx{0.4859} & \xx{0.5723} & \xx{0.4509} & \xx{0.6173} & \xx{0.4506} \\ 
  \hlineB{3}
\end{tabular}
\end{center}
\end{table*}

As described in Section~\ref{sec:task}, the goal of each task is to find the relevant concept in the taxonomy for the given query. Therefore, obtaining at least one surface form associated with the relevant concept at the top of the ranking is sufficient to achieve this goal. With that in mind, we use mean reciprocal rank (MRR) and top-$k$ accuracy (A@$k$) as evaluation metrics.

Due to space constraints, in Table~\ref{tab:results-main-mrr} we present results in terms of mean reciprocal rank (MRR) for a selected subset of tasks, while the complete set of results is provided in Table~\ref{tab:results-appendix-mrr} and Table~\ref{tab:results-appendix-mrr-bis} in Appendix~\ref{sec:appendix-full-results}.


In most monolingual datasets, the top-performing lexical baselines achieved MRR values ranging from 30\% to 55\%. Notably, in the French\footnote{Results for every dataset are presented in Appendix~\ref{sec:appendix-full-results}.} and Danish datasets, these baselines performed extraordinarily well in large part due to substantial lexical overlap, as indicated by the left-skewed distributions in Figure~\ref{fig:hist-alt-all}. In contrast, the Lithuanian, Norwegian, and Romanian datasets exhibited lower performance. Char-based TF-IDF variants deliver the highest performance among this group of baselines.

In a zero-shot setup, ESCOXLM-R performs poorly, even falling behind simple lexical baselines across both monolingual and cross-lingual datasets. This result is consistent with previous research that has shown that encoders trained with masked language modeling (MLM) objectives often struggle to produce effective sentence representations when directly evaluated as sentence encoders~\cite{li-etal-2020-sentence}. In contrast, the other bi-encoders evaluated in this study were specifically optimized for generating useful sentence embeddings, which explains their superior performance in these tasks.

\begin{figure}[t]
\begin{center}

    \subfloat[Absolute performance (in MRR).]
    {
    \label{fig:corr-models-abs}
    \includegraphics[width=0.45\textwidth]{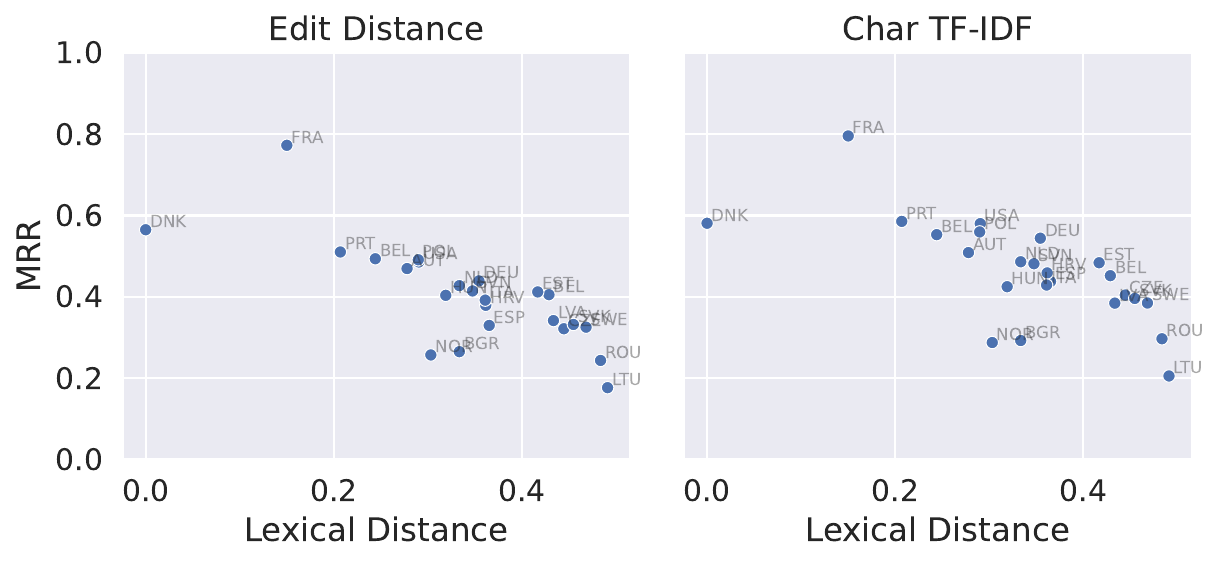}
    } \\

    \par\bigskip

    \subfloat[Performance relative to the lexical baseline Char TF-IDF]
    {
    \label{fig:corr-models-delta}
    \includegraphics[width=0.45\textwidth]{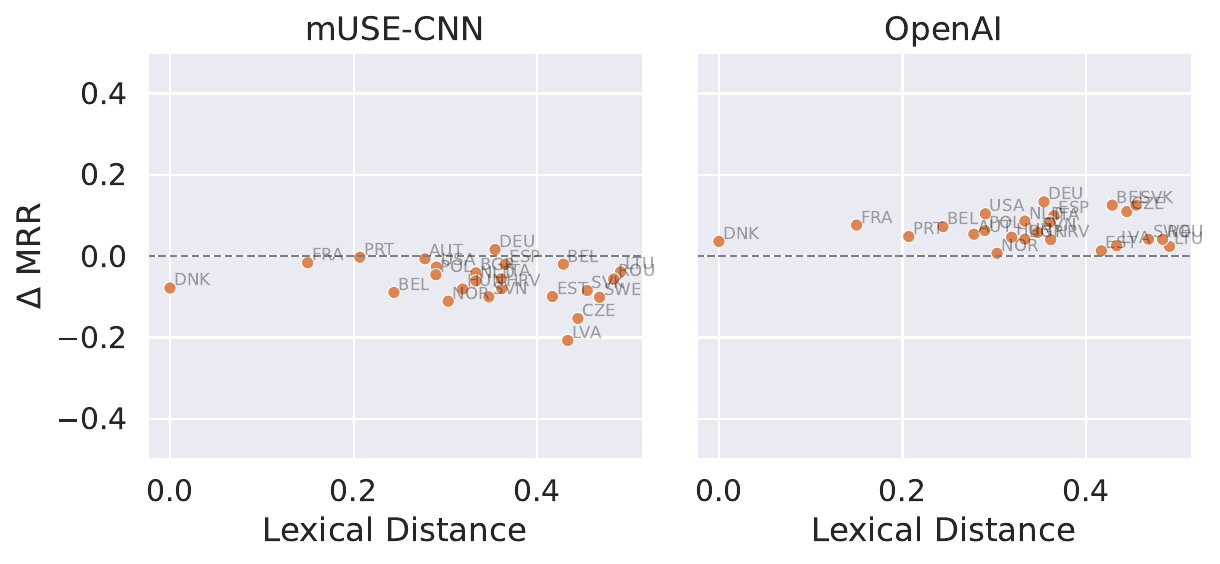}
    } \\

    \caption{Correlation between model performance and the median of the minimum edit distance between queries and relevant corpus elements in monolingual datasets.}

\label{fig:corr-models}
\end{center}
\end{figure}

The mUSE-CNN model demonstrates fair performance on most monolingual tasks for languages included in its pre-training, especially when considering its relatively small model size and architecture type (see Table~\ref{tab:models-detail}). However, as anticipated, its performance drops significantly for languages that were not included during its pre-training. Furthermore, its performance falls below the lexical baselines in almost all datasets. This can be observed in Figure~\ref{fig:corr-models-delta}.

MPNet exhibits poor performance across all monolingual datasets, a surprising result given its larger model size, architecture type, and the fact that it was pre-trained in all the languages used in this experiment. Despite these advantages, it is generally outperformed by the smaller mUSE-CNN model, with the notable exception of the English datasets.

BGE-M3 and Multilingual E5 have similar characteristics, as described in Table~\ref{tab:models-detail}, and both deliver strong performance across most monolingual tasks. In these cases, they generally outperform all lexical baselines and smaller bi-encoders. However, in the English datasets, Multilingual E5 outperforms BGE-M3.

GIST-Embedding demonstrates strong performance in English, outperforming many larger models. It also achieves reasonable results in most other languages, which is surprising considering its primary training was focused on English.

E5, a significantly larger Decoder-only model, outperforms the previously mentioned models across most tasks. This is also surprising since E5 was mainly trained in English. Finally, although limited details are available publicly about OpenAI's \texttt{text-embedding-3-large} model, its performance is generally on par with or even surpasses that of E5. OpenAI's model delivers the highest overall performance among all the models evaluated in our experiments.

The performance of the models in each monolingual dataset is correlated with the lexical overlap in the dataset, as measured by the median of the distributions presented in Figure~\ref{fig:hist-alt-all}. As expected, lexical baselines exhibit a particularly strong correlation, with Spearman's coefficients of -0.74 for Char~TF-IDF and -0.80 for Edit Distance. Interestingly, bi-encoders also demonstrate a moderate correlation, such as mE5~(-0.65) and OpenAI~(-0.62). In Figure~\ref{fig:corr-models} we visualize this correlation, as well as the correlation between the lexical overlap and the difference in performance between some bi-encoders and a lexical baseline\footnote{Same figure is displayed in full size in Appendix~\ref{sec:appendix-full-results}}. We observe that, the less lexical overlap in the dataset, the more the OpenAI model outperforms the lexical baseline.


Comparing the results of datasets USA-en-en and USA-en-xx, which share the same queries, we observe that most methods significantly enhance their performance when the corpus elements visible to the system are expanded to include multiple languages, surpassing their performance in the monolingual task. An implication for this is that, when linking mentions into a multilingual taxonomy, the surface forms in other languages are valuable even if the taxonomy includes entity names in the language of the query.


As expected, the performance drop when moving from monolingual to cross-lingual datasets (excluding O*NET) is significantly more pronounced for the lexical baselines compared to the bi-encoders. The capacity for (zero-shot) cross-lingual EL of occupations varies for different models: ESCOXLM-R, MPNet, and GIST-Embedding exhibit very low cross-lingual performance; mUSE-CNN, BGE-M3, and Multilingual E5 demonstrate fair cross-lingual performance; while E5 and OpenAI achieve the highest cross-lingual performance.

\begin{figure*}[t]
\begin{center}
\includegraphics[width=0.92\textwidth]{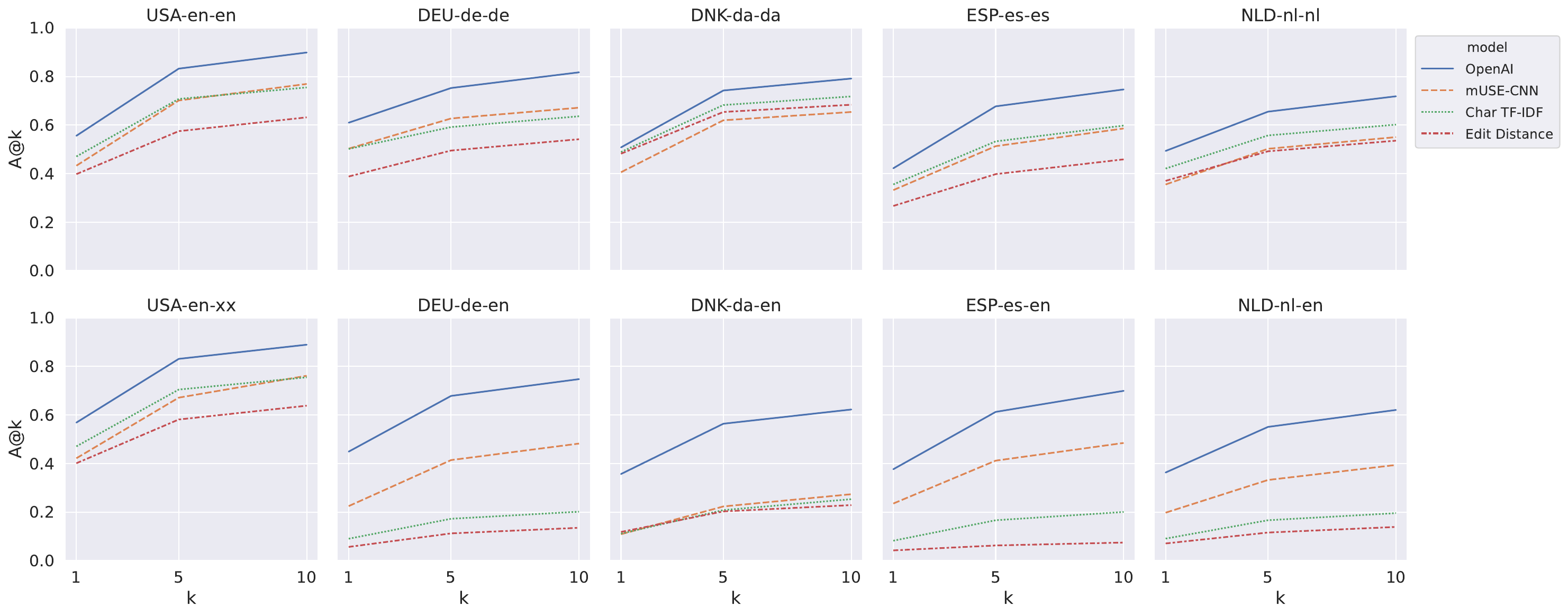}
\caption{Top-$k$ accuracy (A@k) for a selection of models in the MELO Benchmark tasks corresponding to O*NET, Germany, Spain, the Netherlands, and Denmark.}
\label{fig:res-main-acc}
\end{center}
\end{figure*}

Since the techniques we experiment with ---lexical scorers and bi-encoders--- are commonly used for candidate generation in the first stage of EL~\cite{logeswaran-etal-2019-zero, wu-etal-2020-scalable}, it is interesting to measure the top-$k$ accuracy (A@$k$) for different values of $k$ to assess how well such techniques recover the first relevant item. Figure~\ref{fig:res-main-acc} presents these results for the same subset of tasks for the following systems: Edit Distance, Char-level TF-IDF, mUSE-CNN, and OpenAI. The complete set of A@$k$ is available in Appendix~\ref{sec:appendix-full-results}, in Figure~\ref{fig:res-alt-1-acc} and Figure~\ref{fig:res-alt-2-acc}. The results observed for top-$k$ accuracy are consistent with those for mean reciprocal rank (MRR), particularly in terms of the relative ranking and comparative performance of the models.


\section{Related Work}
\label{sec:related-work}

There has been significant research interest in systems that normalize HR information into ESCO and other taxonomies.

Decorte~et~al.~\cite{decorte2023negsampl} explore the extraction of ESCO skills from segmented job descriptions. They approach this problem as a massive multi-label classification task, and present a human-annotated evaluation set for this task. More recently, Decorte~et~al.~\cite{decorte2023skillsgpt} approach the same problem from an EL perspective. They use a large language model~(LLM) to produce synthetic annotations and train a bi-encoder to extract ESCO skills from job description segments. Finally, Zhang~et~al.~\cite{zhang-etal-2024-entity} apply and compare two supervised EL methods for solving the same task: BLINK~\cite{wu-etal-2020-scalable} and GENRE~\cite{cao2021autoregressive}. In contrast to these other studies, our work focuses on occupations instead of skills, explores cross-lingual and multilingual scenarios, 
and the task as we formulate it does not use context for linking the query mentions.

There has also been a substantial amount of research focused on occupations. Decorte~et~al.~\cite{decorte2021} developed an unsupervised approach to fine-tune BERT~\cite{devlin-etal-2019-bert} to encode the semantics of occupation names. Furthermore, they create a dataset for the normalization of free-form English occupation names into ESCO and they use it to evaluate their model. It has been reported that this dataset contains ambiguous input queries~\cite{decorte2021} as well as some mislabeled elements~\cite{escoML}. Closely related works by Zbib~et~al.~\cite{zbib2022jobtitlesim} and by Bocharova~et~al.~\cite{bocharova2023vacancysbert} propose alternative unsupervised representation learning schemes. They both release evaluation datasets, the former for occupation name ranking, and the latter for EL of unnormalized occupation names into ESCO.

Lake~\cite{lake2022jobclass} studies the application of bi-encoders and cross-encoders to EL of occupations to a custom taxonomy. Yamashita~et~al.~\cite{yamashita2023james} work on a normalization task for occupations, which closely resembles our formulation of EL. They create a non-public dataset by collecting a large number of unnormalized occupation names and then automatically mapping them to ESCO occupations via exact match after removing proper nouns. Vrolijk~et~al.~\cite{vrolijk2023enhancing} build a synthetic dataset for zero-shot evaluation and fine-tuning of several language models using information from ESCO that includes the synonyms for each entity name, the relationship between entities, and their definitions. In particular, they use the set of name synonyms for each ESCO occupation to pose a binary relevance classification problem, where positive pairs involve two names belonging to the same synonym set.

Two important use cases of the EL task under study are enriching and aligning taxonomies. In order to maintain up-to-date but well-curated taxonomies, it is common to automatically identify new candidate concepts to be included, and to use human annotators to validate their inclusion. Similarly, when aligning two taxonomies ---i.e.~building a crosswalk---, it is common to use automatic systems to propose and explore candidate matches between the concepts in each taxonomy.

Giabelli~\cite{giabelli2024phdthesis} and colleagues have worked on several approaches for enriching~\cite{giabelli2020neo} and aligning~\cite{giabelli2021weta, giabelli2022jota} taxonomies using word embeddings to model concepts via their names, together with structural information about the taxonomy. All these methods automatically score candidates for inclusion or mapping, and can be used within a human-in-the-loop framework for further validation.

During the creation of the crosswalk between O*NET and ESCO, the teams responsible for maintaining both taxonomies worked together to ensure a high-quality mapping~\cite{escoOnetCrosswalk}. Interestingly, they report employing a human-in-the-loop methodology where a fine-tuned BERT model~\cite{devlin-etal-2019-bert} is used as a bi-encoder to rank the ESCO occupations for each O*NET occupation. They explore different methods for encoding each, leveraging occupation names (and synonyms) as well.

More recently, the ESCO team presented an analysis~\cite{escoML} on a task that is very similar to the one we present here. They fine-tune a XLM-RoBERTa model~\cite{conneau-etal-2020-unsupervised} on HR-related data, including the textual information from ESCO, but with no supervision signal for any specific EL task. They then use this model as a bi-encoder to suggest ESCO occupations for elements taken from the national terminologies of Latvia, Spain, Sweden, and Italy, as well as from O*NET. Using the respective crosswalks, they evaluate this as an EL task. They explore monolingual and cross-lingual (to English) modalities. A key difference between this work and ours is that they consider any SKOS relationship as a legitimate annotation, while we only use exact and narrow matches. We also filter out semantically ambiguous queries for which experts determined that they should be related as an exact match to more than one ESCO concept. For those reasons, their results are not comparable to those we present in this work.


\section{Conclusion}
\label{sec:conclusions}

We have introduced the MELO Benchmark, a suite of 48 datasets for multilingual entity linking of occupations in 21 languages. We experimented with several out-of-the-box lexical and semantic baselines, demonstrating that there is still significant room for improvement. Our aim is that MELO will serve as a valuable resource for the research community, providing a standardized benchmark for assessing progress in multilingual EL within the HR domain, and fostering innovation and the development of new methodologies in this important area of research.

In future work, several research directions could be explored. First, the current evaluation scheme can be extended to incorporate NIL prediction or prediction using entity descriptions rather than relying solely on entity names, with the presented source code being easily adaptable for such modifications. Second, domain-adapting or fine-tuning encoders specifically for this task, in a manner similar to ESCOXLM-R but optimized for semantic text similarity, presents another possible direction. Third, exploring advanced deep learning techniques beyond bi-encoders, such as cross-encoders combined with re-ranking stages, could enhance model performance. Finally, investigating the meta-learning paradigm by dividing MELO tasks into meta-training and meta-testing tasks, and applying meta-learning context to solve the meta-testing tasks, exploiting multi-lingual transfer capabilities of modern deep-learning models, offers another interesting direction for future work.


\section*{Acknowledgment}
\label{sec:ethics}

This publication uses the ESCO classification of the European Commission. We gratefully acknowledge the work done by the team involved in curating the ESCO Occupations taxonomy, as well as the teams responsible for the O*NET-SOC 2019 taxonomy and the other national taxonomies used in this work. Furthermore, we would also like to thank the teams responsible for creating the crosswalks between ESCO and these taxonomies.

\bibliography{main}

\appendix


\section{Details on the Datasets}
\label{sec:appendix-details-datasets}

\begin{figure*}[t]
\begin{center}
\includegraphics[width=0.87\textwidth]{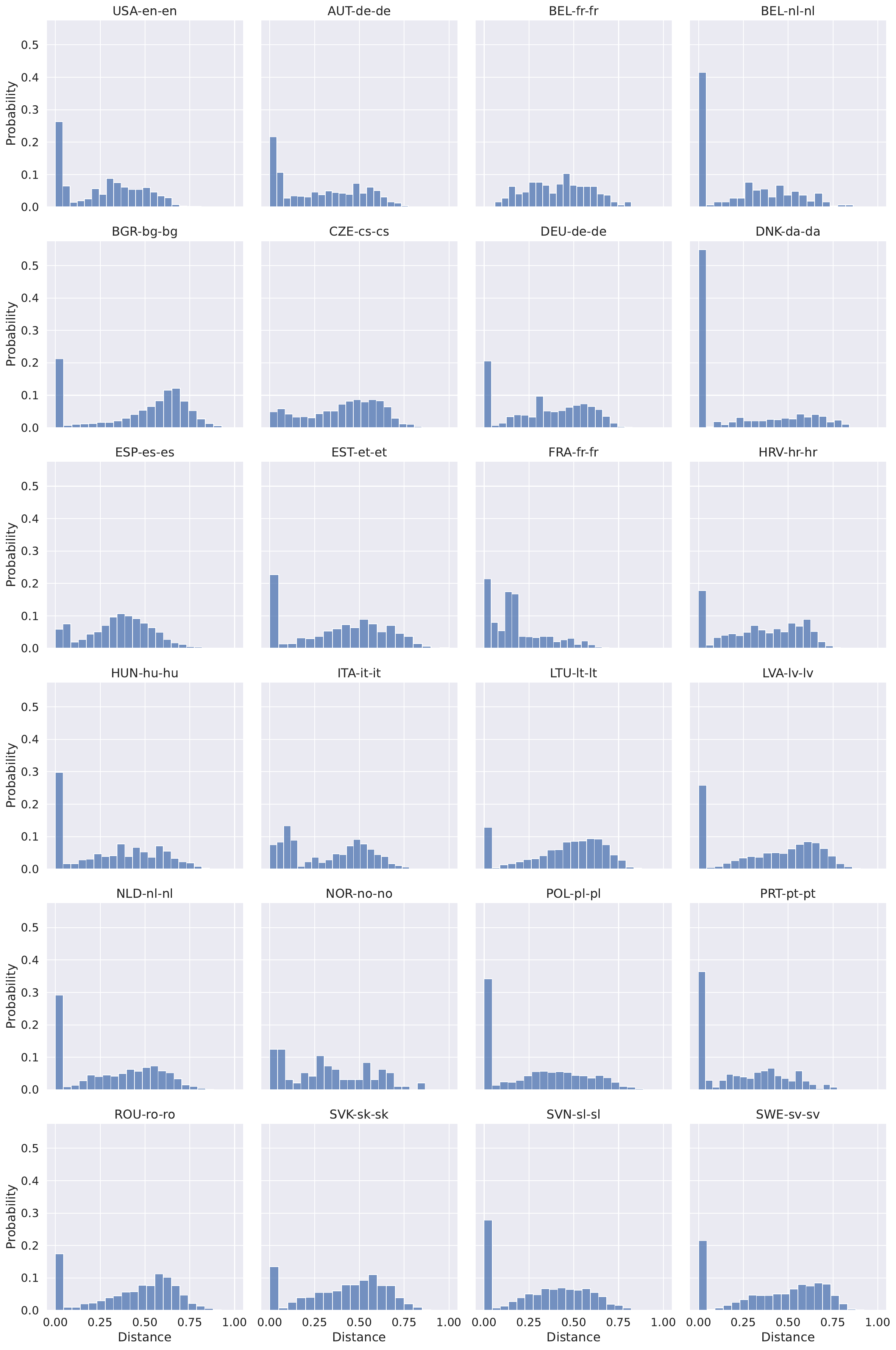}
\caption{Histogram of minimum (normalized) edit distances between each query and the closest relevant corpus element for each monolingual task in MELO.}
\label{fig:hist-alt-all}
\end{center}
\end{figure*}


We release the source code used to build the datasets\footnote{\url{https://github.com/Avature/melo-benchmark}}, providing researchers with a tool to easily generate new datasets by combining different sets of languages for query and corpus elements. Using this code, new instantiations of the task can be derived from the input data by defining custom language combinations. For example, it is possible to use the Italian national terminology to set up an Italian-to-Greek cross-lingual task, or even combine the query sets of several national classifications and leverage all languages in ESCO to create a more complex multilingual task.

The input data consists of files with the multilingual ESCO Occupations taxonomy (one for each relevant version) and files containing the queries in each national terminology, which are mapped to the ESCO concept ID of the relevant occupation. To create a dataset, the user can select a national terminology and a set of languages for the corpus (any subset of the languages supported by ESCO). 

In Table~\ref{tab:dataset-examples} we present example queries and their relevant corpus elements, sampled from the NLD-nl-nl, PRT-pt-pt, and PRT-pt-en datasets.

Finally, we analyze the lexical overlap between the national classifications and ESCO. In Figure~\ref{fig:hist-alt-all}, we present a histogram showing the normalized edit distance between queries and their closest relevant corpus element, for all the tasks in MELO.

To compute the distances, we first lowercase the surface forms of both the query and the corpus element, and we use the method \texttt{ratio} from the Python package \textsc{rapidfuzz}\footnote{\url{https://rapidfuzz.github.io/RapidFuzz/Usage/fuzz.html}}. This is a measure of the normalized edit distance between the two strings. In the histograms, for each query, we compute the distance for all its relevant corpus elements and report the minimum distance.

In the histograms, the left-most bin represents the fraction of queries for which the closest relevant element is either identical or very similar. The Danish national terminology has the highest concentration of such cases. To a lesser extent, this is also true for Hungarian, Estonian, and Polish.

Excluding those lexically trivial cases, the more the distribution is skewed to the left, the easier the task. For example, comparing the Belgian (in the French language) and the French tasks, the queries from the French terminology show greater lexical overlap with their relevant corpus elements.

In Appendix~\ref{sec:appendix-full-results}, we use this analysis to compare the performance of lexical baselines across different monolingual tasks.


\begin{table*}[htbp]
\footnotesize
\begin{center}
{\caption{
Characteristics of models used in bi-encoder experiments. Encoder-only and decoder-only architectures refer to Transformers. Model size is given in millions of parameters. Some specifications are unknown for the OpenAI model. The \textit{Language Support} column indicates the extent to which the languages involved in the benchmark are supported by each model.\ \textsuperscript{\color{mybluex} \textdagger}~The mUSE-CNN model supports only English, German, French, Spanish, Dutch, Portuguese, Italian, and Polish.\ \textsuperscript{\color{mybluex} \ddag}~Although GIST-Embedding and E5 are reported to be trained primarily in English, the pre-training of these models did involve examples in other languages as well.
}\label{tab:models-detail}}
\begin{tabular}{llrrl}
  \hlineB{3}
  Model                &  Architecture              & Model Size & Output Dims & Language Support \\
  \hlineB{3}

  ESCOXLM-R            &  Encoder-only Transformer  & 561        & 1024        & Complete        \\ 
  mUSE-CNN             &  CNN                       & 69         & 512         & Partial \textsuperscript{\color{mybluex} \textdagger}   \\ 
  Paraph-mMPNet        &  Encoder-only Transformer  & 278        & 768         & Complete        \\ 
  BGE-M3               &  Encoder-only Transformer  & 560        & 1024        & Complete        \\ 
  GIST-Embedding       &  Encoder-only Transformer  & 109        & 768         & Mainly English \textsuperscript{\color{mybluex} \ddag} \\ 
  mE5                  &  Encoder-only Transformer  & 560        & 1024        & Complete        \\ 
  E5                   &  Decoder-only Transformer  & 7111       & 4096        & Mainly English \textsuperscript{\color{mybluex} \ddag} \\ 
  OpenAI               &  Unknown                   & Unknown    & 3072        & Unknown         \\ 

  \hlineB{3}
\end{tabular}
\end{center}
\end{table*}

\section{Details on the Models}
\label{sec:appendix-details-models}

Here, we provide further details about the models explored in this work. 

Regarding the lexical baselines, we always apply a simple preprocessing in which we lowercase the input strings and, for all languages except Bulgarian, also perform ASCII normalization. For the edit distance baseline, we use \textsc{rapidfuzz} as described above. For the TF-IDF baselines, we use the \textsc{scikit-learn}\footnote{\url{https://scikit-learn.org/stable/modules/generated/sklearn.feature_extraction.text.TfidfVectorizer.html}} Python package, while for the BM5 variants, we use the Okapi BM25 implementation from \textsc{rank-bm25}\footnote{\url{https://pypi.org/project/rank-bm25/}}.

For the baseline variants that involve lemmatization, we use \textsc{spacy}\footnote{\url{https://spacy.io/api/lemmatizer}} models whenever available. However, \textsc{spacy} models were not available for the following languages: Bulgarian, Czech, Estonian, Hungarian, Latvian, and Slovak. Lemmatization is applied before ASCII normalization.

In the case of bi-encoders, we experiment with several deep learning sentence encoders that have demonstrated strong performance in other semantic text similarity tasks.

The first model is ESCOXLM-R, proposed by Zhang~et~al.~\cite{zhang-etal-2023-escoxlm}, which is based on XLM-RoBERTa. We use the PyTorch implementation and the pre-trained weights that are available on HuggingFace with the model name \texttt{jjzha/esco-xlm-roberta-large}. The base model was pre-trained on data in 88 languages, including all those involved in our datasets, and the fine-tuning by Zhang and colleagues involved learning objectives that leverage information in ESCO. Although it is usual to experiment with the XLM-RoBERTa family of models only after fine-tuning, in our experiment we use it out-of-the-box in a zero-shot setup. During inference, the input to the model is the surface form of the query or the corpus element, with no preprocessing.

We also present results for the Multilingual Universal Sentence Encoder (mUSE-CNN) model variant with a CNN architecture, proposed by Cer~et~al.~\cite{cer-etal-2018-universal, yang-etal-2020-multilingual}. In our experiments, we use the TensorFlow implementation and the pre-trained weights available on TensorFlow Hub with the handle \texttt{google/universal-sentence-encoder-multilingual/3}. This model was pre-trained on data in Arabic, Chinese, English, French, German, Italian, Japanese, Korean, Dutch, Polish, Portuguese, Spanish, Thai, Turkish, and Russian. (Note that, during training, mUSE-CNN has not seen text for languages such as Bulgarian, Czech, or Danish.) During inference, the input to the model is the surface form of the query or the corpus element without any preprocessing or enclosing prompt template.

Other open-source models we experiment with are implemented in PyTorch within the HuggingFace package \textsc{sentence-transformers}~\cite{reimers-gurevych-2019-sentence}. These models are the following: a multilingual model based on MPNet~\cite{song2020mpnet} that was pre-trained on 50 languages, including all of MELO languages\footnote{\url{https://huggingface.co/sentence-transformers/paraphrase-multilingual-mpnet-base-v2}}; the BGE-M3 model~\cite{chen2024bge}, which supports more than 100 languages, including also all MELO languages\footnote{\url{https://huggingface.co/BAAI/bge-m3}}; GIST Embedding \cite{solatorio2024gistembed}, which is a model reported to be primarily trained in English\footnote{\url{https://huggingface.co/avsolatorio/GIST-Embedding-v0}}; Multilingual E5 \cite{wang2024multilingual}, which was pre-trained on 94 languages, including all of MELO languages\footnote{\url{https://huggingface.co/intfloat/multilingual-e5-large}}; and E5 \cite{wang2023improving, wang2022text} pre-trained on many languages but reported to perform best on English-language input\footnote{\url{https://huggingface.co/intfloat/e5-mistral-7b-instruct}}.

Finally, we also experiment with the \texttt{text-embedding-3-large} model from OpenAI\footnote{\url{https://openai.com/index/new-embedding-models-and-api-updates}}, which is reported to be state-of-the-art for many semantic text similarity tasks.

For HuggingFace and OpenAI models, during inference, we wrap the input text (the surface form of the query or corpus element) with the following prompt template:

\begin{tcolorbox}[colback=gray!10!white, colframe=gray!95, arc=2mm, boxrule=0.5mm]
\begin{small}
\begin{center}
\ttfamily
The candidate's job title is \quotes{\{\{surf\_form\}\}}.
What skills are likely required for this job?
\end{center}
\end{small}
\end{tcolorbox}

\noindent where \texttt{\{\{surf\_form\}\}} is replaced with the surface form of the element that is being encoded.

This decision was informed by preliminary experiments in which we evaluated various models with different wrapping prompt templates, including no template (as with ESCOXLM-R and mUSE-CNN). We speculate that such prompts are particularly beneficial for LLM-based encoders, as they may better capture the semantics of the occupation names we aim to rank.

Although we also experimented with prompts in the same language as each query, this did not improve performance. Consistently using a single prompt ensures a language-agnostic and symmetric bi-encoder approach.


\begin{table*}[t]
\footnotesize
\begin{center}
{
\caption{
Example queries and their relevant corpus elements for various datasets in MELO. Surface forms in Dutch are presented in \textcolor{myredb}{red}, while those in Portuguese are in \textcolor{myblueb}{blue}.
}
\label{tab:dataset-examples}
}
\begin{tabular}{lp{5cm}p{6cm}}
  \hlineB{3}
  Dataset Name & Query & Relevant Corpus Elements \\ 
  \hlineB{3}
  \multirow{3}{*}{NLD-\textcolor{myredb}{nl}-\textcolor{myredb}{nl}} 
    & \textcolor{myredb}{
        Kredietbeoordelaar
    } & \textcolor{myredb}{
        kredietanalist
        \newline
        kredietadviseur
        \newline
        analist kredieten en risico's
    } \\ \cline{2-3} 
    & \textcolor{myredb}{
        Woonbegeleider gezinsvervangend huis, wooncentrum
    } & \textcolor{myredb}{
        medewerker verzorgingshuis
        \newline
        medewerker verzorgingstehuis
        \newline
        medewerkster verzorgingscentrum
        \newline
        medewerkster verzorgingstehuis
        \newline
        medewerkster verzorgingshuis
        \newline
        medewerker verzorgingscentrum
    } \\ \cline{2-3} 
    & \textcolor{myredb}{
        PR-adviseur
    } & \textcolor{myredb}{
        lobbyist
        \newline
        lobbyiste
    } \\ 
  \hlineB{3}
  \multirow{3}{*}{PRT-\textcolor{myblueb}{pt}-\textcolor{myblueb}{pt}} 
    & \textcolor{myblueb}{
        Guia intérprete
    } & \textcolor{myblueb}{
        Guias-intérpretes
    } \\ \cline{2-3} 
    & \textcolor{myblueb}{
        Fumigador e outros controladores, de pragas e ervas daninhas
    } & \textcolor{myblueb}{
        Pulverizador de pesticidas/Pulverizadora de pesticidas
        \newline
        Pulverizador de pesticidas
        \newline
        Pulverizadora de pesticidas
    } \\ \cline{2-3} 
    & \textcolor{myblueb}{
        Empregado de serviços de apoio à produção
    } & \textcolor{myblueb}{
        Coordenador de montagem de máquinas/Coordenadora de montagem de máquinas
        \newline
        Coordenador de montagem de máquinas
        \newline
        Coordenadora de montagem de máquinas
    } \\ 
  \hlineB{3}
  \multirow{3}{*}{PRT-\textcolor{myblueb}{pt}-en}
    & \textcolor{myblueb}{
        Guia intérprete
    } & 
        Travel guides
    \\ \cline{2-3} 
    & \textcolor{myblueb}{
        Fumigador e outros controladores, de pragas e ervas daninhas
    } & 
        pesticides sprayer
        \newline
        lawn care chemical applicator
        \newline
        spray technician
        \newline
        pesticides applicator
        \newline
        trees sprayer
        \newline
        sprayer of pesticides
    \\ \cline{2-3} 
    & \textcolor{myblueb}{
        Empregado de serviços de apoio à produção
    } & 
        machinery assembly coordinator
        \newline
        production line coordinator
        \newline
        manufacturing co-ordinator
        \newline
        assembly line coordinator
        \newline
        machinery manufacturing co-ordinator
        \newline
        machinery production inspector
        \newline
        machinery production co-ordinator
        \newline
        assembly line co-ordinator
        \newline
        machinery assembly co-ordinator
        \newline
        machinery manufacturing manager
        \newline
        production line co-ordinator
    \\ 
  \hlineB{3}
\end{tabular}
\end{center}
\end{table*}


\section{Full Results}
\label{sec:appendix-full-results}

This Section presents the full set of experimental results. Table~\ref{tab:results-appendix-mrr} and Table~\ref{tab:results-appendix-mrr-bis} include the mean reciprocal rank (MRR) for each model across all tasks in MELO.

Although not included with the main results, we also evaluated a random baseline for each dataset, where the score $s(q, c_{i})$ for any query and any corpus element is drawn from a uniform distribution. The performance of this baseline varies depending on the number of corpus elements and the distribution of relevant elements per query, but in general, its MRR is close to 0.020.

\begin{figure*}[t]
\begin{center}

    \subfloat[Absolute performance (in MRR).]
    {
    \label{fig:corr-models-big-abs}
    \includegraphics[width=0.82\textwidth]{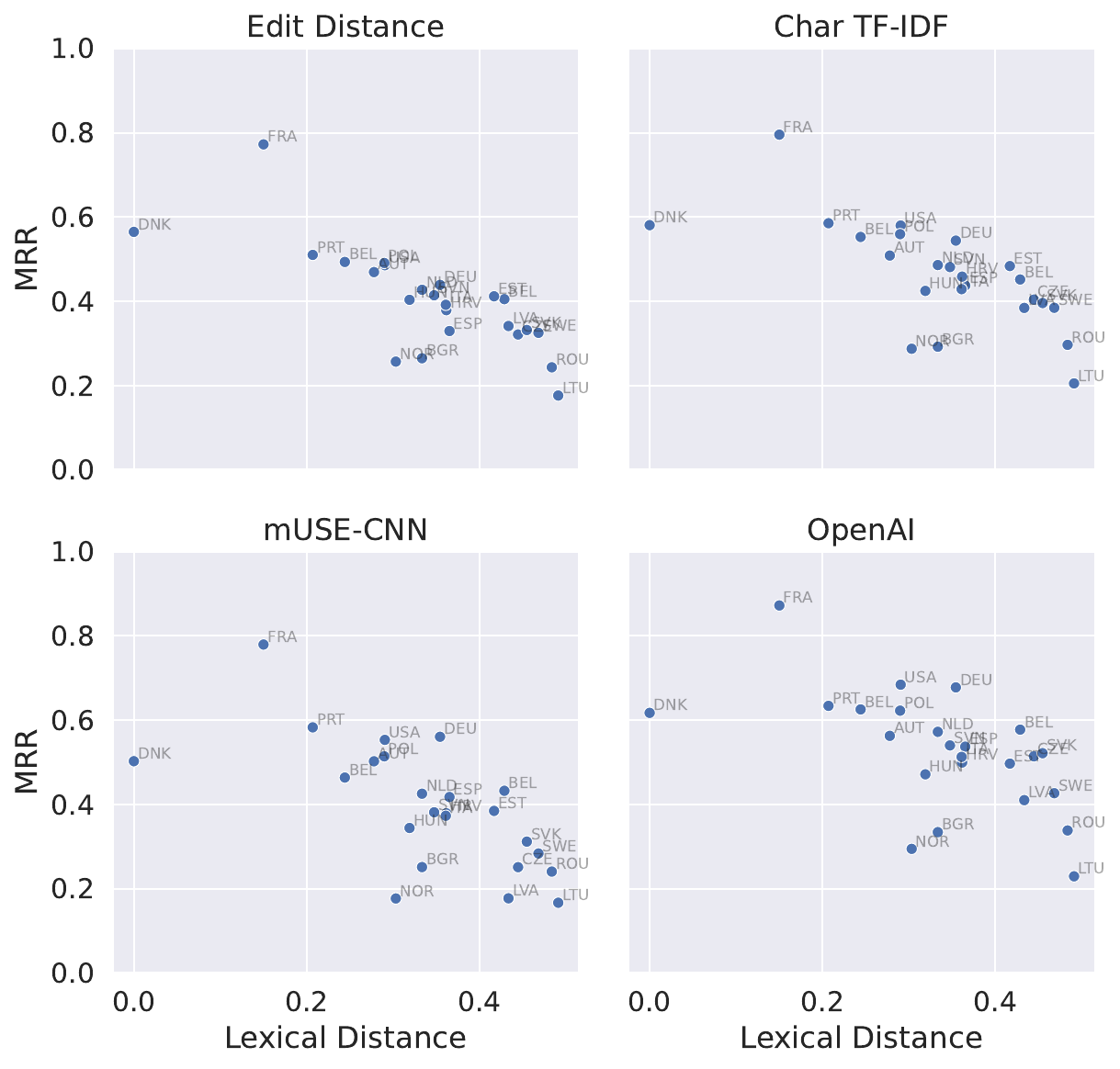}
    } \\

    \par\bigskip

    \subfloat[Performance relative to the lexical baseline Char TF-IDF]
    {
    \label{fig:corr-models-big-delta}
    \includegraphics[width=0.82\textwidth]{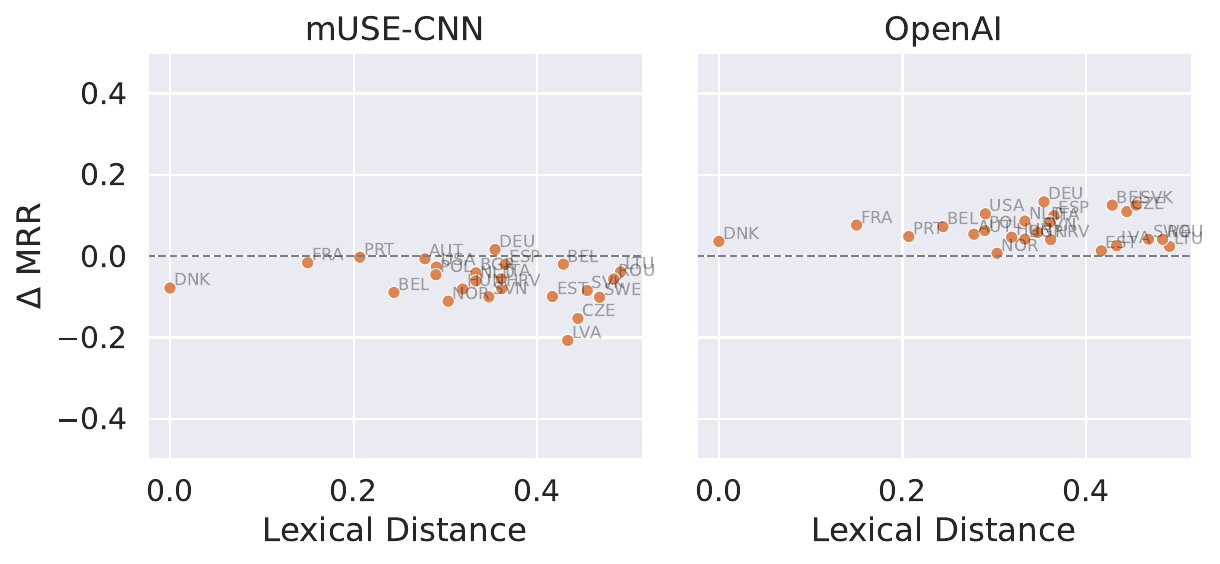}
    } \\

    \caption{Correlation between model performance and the median of the minimum edit distance between queries and relevant corpus elements in monolingual datasets.}

\label{fig:corr-models-big}
\end{center}
\end{figure*}

Additionally, Figure~\ref{fig:corr-models-big} shows scatterplots illustrating the correlation between model performance and the median of the lexical overlap index described in Appendix~\ref{sec:appendix-details-datasets}: the minimum normalized edit distance per query.

\begin{figure*}[t]
\begin{center}

    \subfloat[Results for tasks: O*NET, Austria, Belgium (fr), Belgium (nl), and Bulgaria.]
    {
    \label{fig:res-alt-1-acc-a}
    \includegraphics[width=0.92\textwidth]{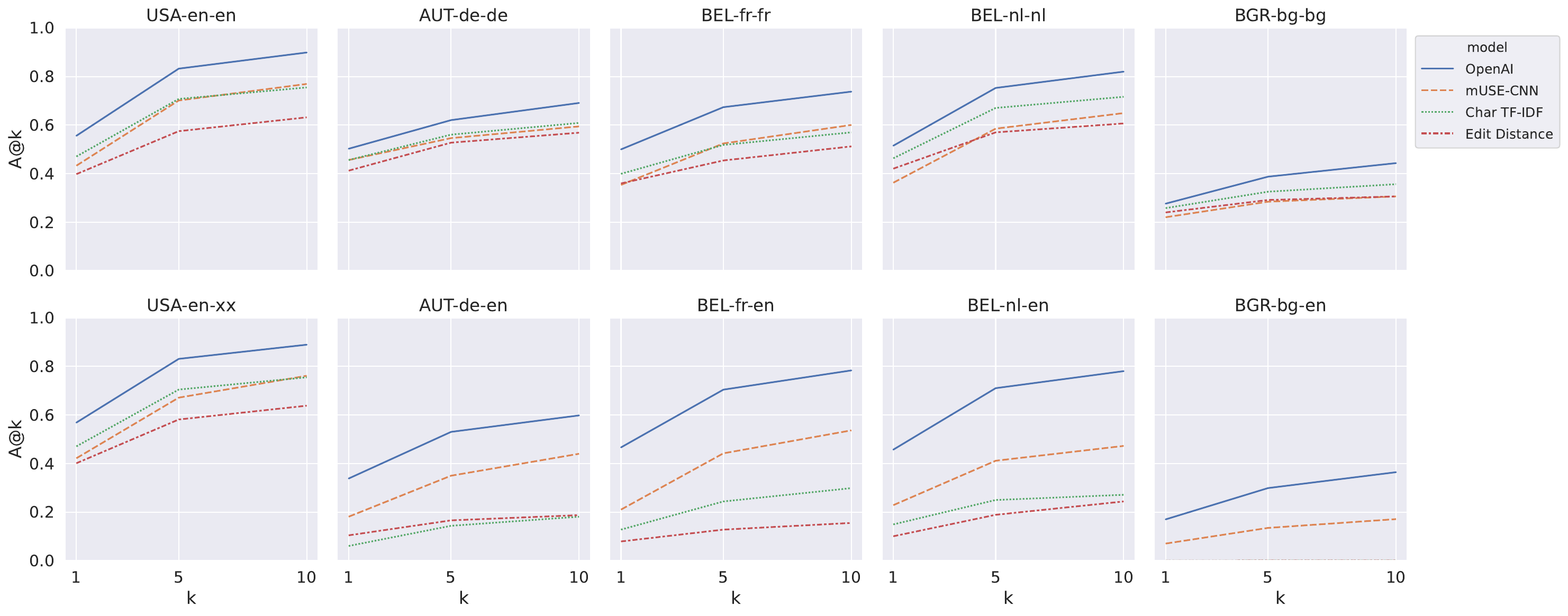}
    } \\

    \par\bigskip

    \subfloat[Results for tasks: Czechia, Germany, Denmark, Spain, and Estonia.]
    {
    \label{fig:res-alt-1-acc-b}
    \includegraphics[width=0.92\textwidth]{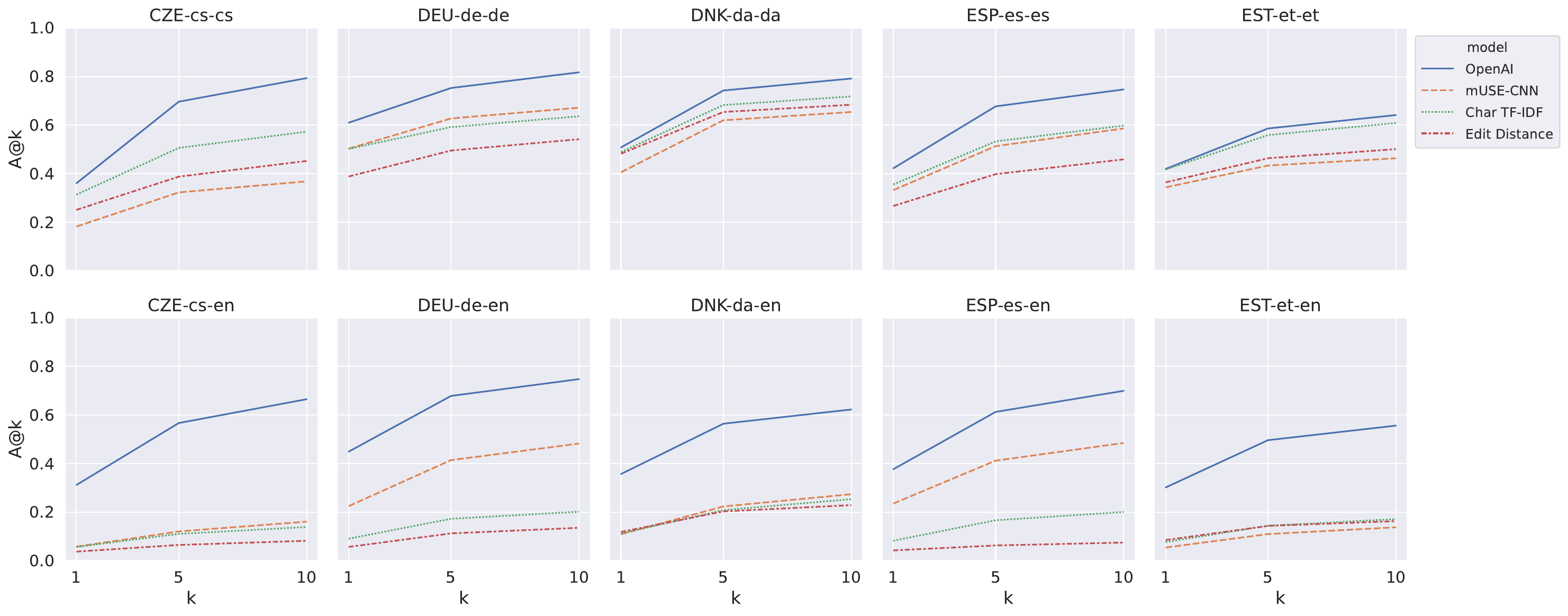}
    } \\

    \par\bigskip

    \subfloat[Results for tasks: France, Croatia, Hungary, Italy, and Lithuania.]
    {
    \label{fig:res-alt-1-acc-c}
    \includegraphics[width=0.92\textwidth]{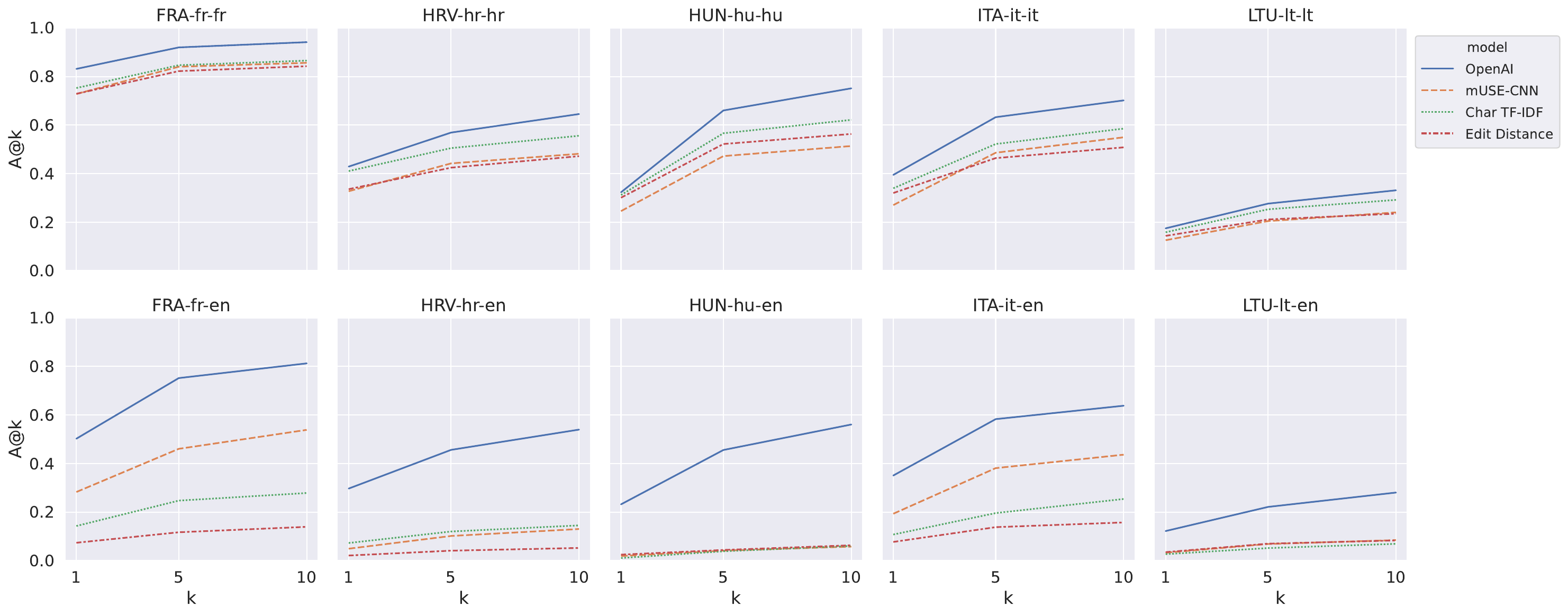}
    }

    \caption{Top-$k$ accuracy (A@k) for a selection of models in the MELO Benchmark tasks.}

\label{fig:res-alt-1-acc}
\end{center}
\end{figure*}

\begin{figure*}[t]
\begin{center}

    \subfloat[Results for tasks: Latvia, the Netherlands, Norway, Poland, and Portugal.]
    {
    \label{fig:res-alt-2-acc-a}
    \includegraphics[width=0.92\textwidth]{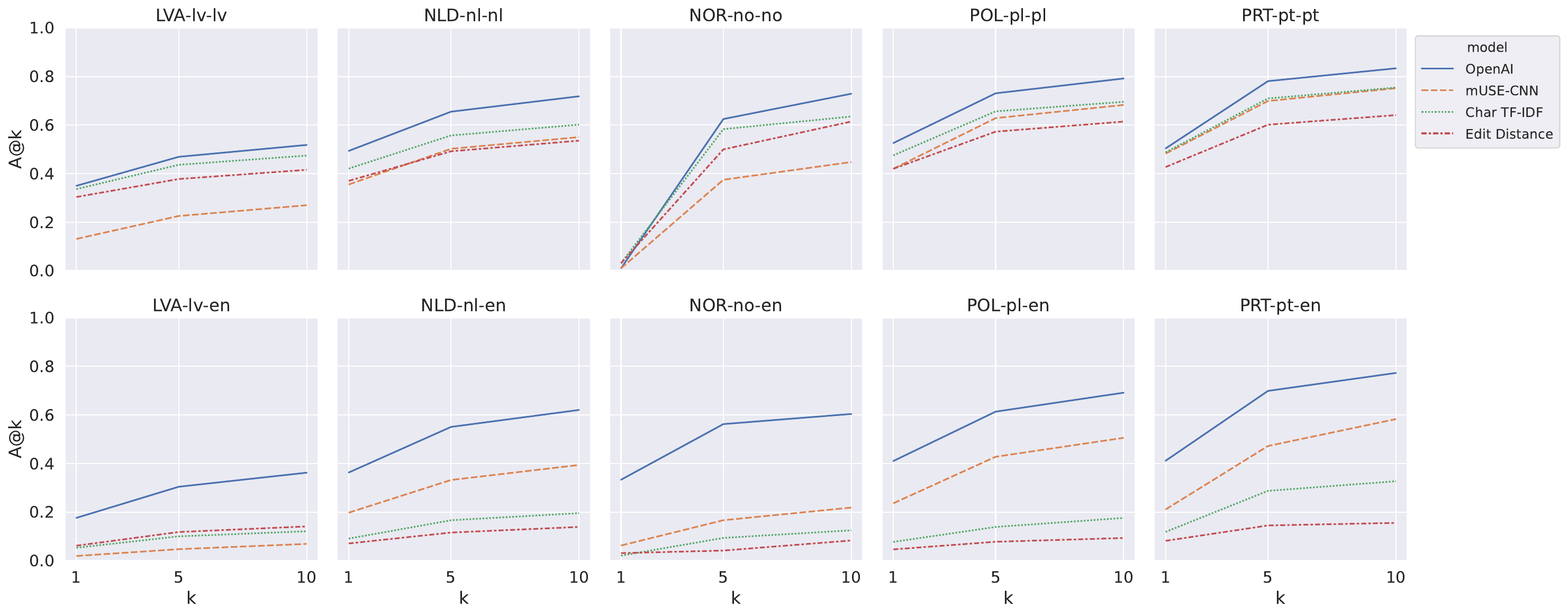}
    } \\

    \par\bigskip

    \subfloat[Results for tasks: Romania, Slovakia, Slovenia, and Sweden.]
    {
    \label{fig:res-alt-2-acc-b}
    \includegraphics[width=0.78\textwidth]{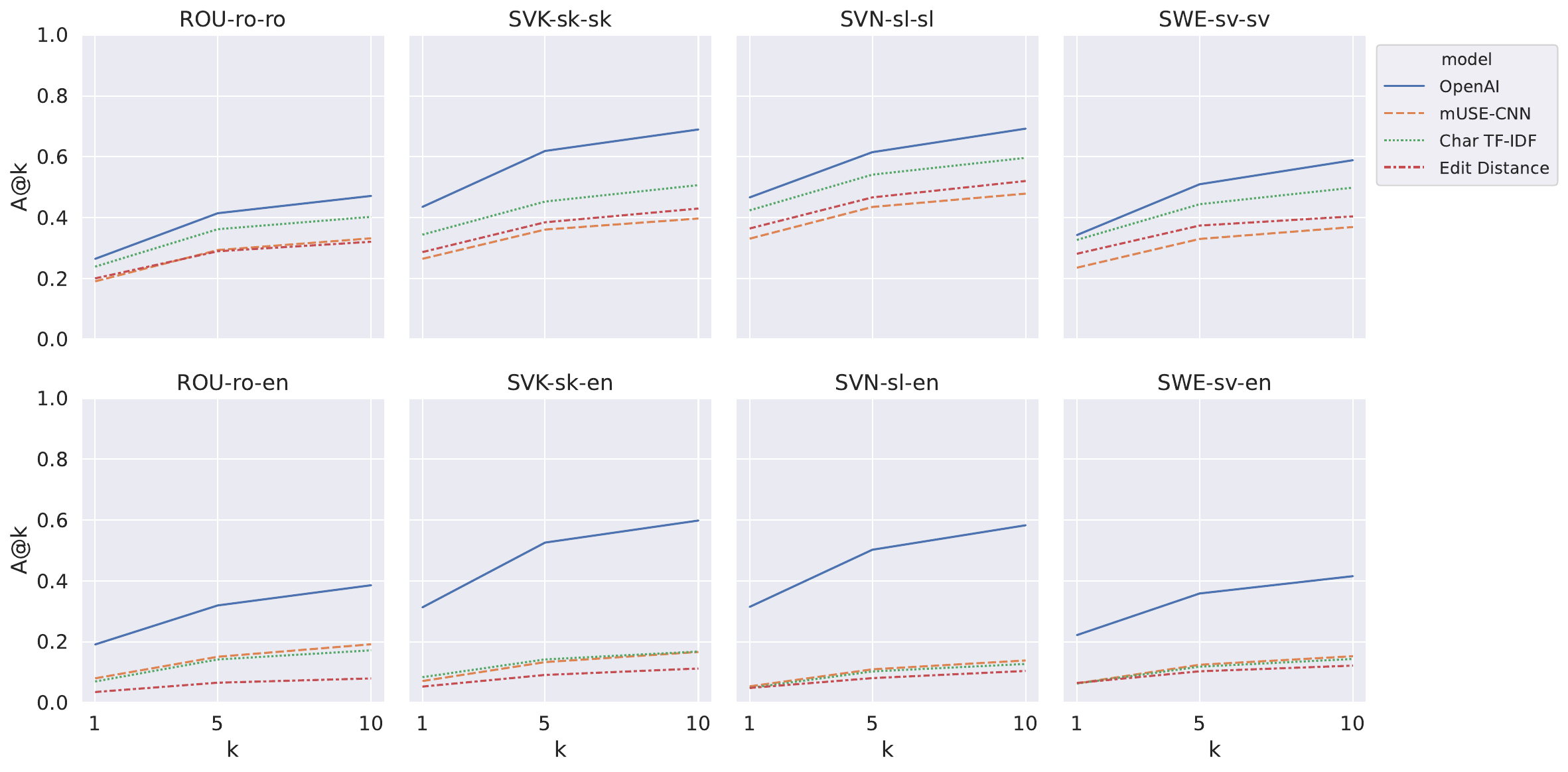}
    } \\

    \caption{Top-$k$ accuracy (A@k) for a selection of models in the MELO Benchmark tasks.}

\label{fig:res-alt-2-acc}
\end{center}
\end{figure*}

Finally, in Figure~\ref{fig:res-alt-1-acc} and in Figure~\ref{fig:res-alt-2-acc} we show the top-$k$ accuracy (A@$k$) for a selection of models in every task in MELO.


\begin{table*}[htbp]
\footnotesize
\begin{center}
    {
    \caption{
    Mean reciprocal rank (MRR) for every model, evaluated in the monolingual and the cross-lingual versions of the MELO tasks.
    }
    \label{tab:results-appendix-mrr}
    }
    \centering

    \begin{subtable}{\textwidth}
    \centering
    \begin{tabular}{lcccccccccc}

        \hlineB{3}
        \multirow{2}{*}{Model} &
        \multicolumn{2}{c}{USA} &
        \multicolumn{2}{c}{AUT} &
        \multicolumn{2}{c}{BEL} &
        \multicolumn{2}{c}{BEL} &
        \multicolumn{2}{c}{BGR}
        \\
        \cline{2-11} 
                        & en-en  & en-xx & de-de & de-en & fr-fr & fr-en & nl-nl & nl-en & bg-bg & bg-en \\ 
        \hlineB{3}
  Edit Distance        & 0.4858      & 0.4889      & 0.4695      & 0.1337      & 0.4053      & 0.1072      & 0.4936      & 0.1456      & 0.2651      & 0.0007      \\ 
  Word TF-IDF          & 0.3250      & 0.3207      & 0.4104      & 0.0319      & 0.4589      & 0.0735      & 0.4914      & 0.0618      & 0.2740      & 0.0033      \\ 
  Word TF-IDF (lemmas) & 0.6056      & 0.5999      & 0.4115      & 0.0288      & 0.4677      & 0.0947      & 0.4907      & 0.0593      &  -          &  -          \\ 
  Char TF-IDF          & 0.5800      & 0.5764      & 0.5088      & 0.1008      & 0.4520      & 0.1827      & 0.5529      & 0.1970      & 0.2925      & 0.0006      \\ 
  Char TF-IDF (lemmas) & 0.5957      & 0.5913      & 0.5096      & 0.1269      & 0.4597      & 0.1781      & 0.5474      & 0.1750      &  -          &  -          \\ 
  BM25                 & 0.2936      & 0.2814      & 0.0252      & 0.0041      & 0.4130      & 0.0583      & 0.4553      & 0.0398      & 0.2581      & 0.0033      \\ 
  BM25 (lemmas)        & 0.6004      & 0.5978      & 0.3808      & 0.0186      & 0.4651      & 0.0723      & 0.4598      & 0.0389      &  -          &  -          \\ 
        \hline
  ESCOXLM-R            & 0.3450      & 0.3426      & 0.4150      & 0.0767      & 0.2575      & 0.0537      & 0.3720      & 0.1084      & 0.2215      & 0.0269      \\ 
  mUSE-CNN             & 0.5532      & 0.5317      & 0.5024      & 0.2656      & 0.4324      & 0.3213      & 0.4638      & 0.3148      & 0.2514      & 0.1044      \\ 
  Paraph-mMPNet        & 0.5876      & 0.5822      & 0.3726      & 0.0852      & 0.3824      & 0.1459      & 0.4283      & 0.1498      & 0.2146      & 0.0167      \\ 
  BGE-M3               & 0.6226      & 0.6301      & 0.5330      & 0.2819      & 0.5225      & 0.4005      & 0.5709      & 0.3529      & 0.3192      & 0.1825      \\ 
  GIST-Embedding       & 0.6431      & 0.6464      & 0.4819      & 0.0947      & 0.4803      & 0.1848      & 0.5113      & 0.1706      & 0.2700      & 0.0033      \\ 
  mE5                  & 0.6563      & 0.6588      & 0.5334      & 0.3092      & 0.5407      & 0.4266      & 0.5683      & 0.3851      & 0.3106      & 0.1870      \\ 
  E5                   & 0.6735      & 0.6777      & 0.5612      & 0.4143      & 0.5606      & 0.5380      & 0.6133      & 0.4991      & 0.3406      & 0.2371      \\ 
  OpenAI               & 0.6842      & 0.6872      & 0.5628      & 0.4304      & 0.5775      & 0.5736      & 0.6255      & 0.5698      & 0.3343      & 0.2367      \\ 
        \hlineB{3}

    \end{tabular}
    \caption{
    Results for tasks: O*NET, Austria, Belgium (French), Belgium (Dutch), and Bulgaria.
    }
    \label{tab:results-appendix-mrr-a}
    \end{subtable}

    \par\bigskip

    \begin{subtable}{\textwidth}
    \centering
    \begin{tabular}{lcccccccccc}

        \hlineB{3}
        \multirow{2}{*}{Model} &
        \multicolumn{2}{c}{CZE} &
        \multicolumn{2}{c}{DEU} &
        \multicolumn{2}{c}{DNK} &
        \multicolumn{2}{c}{ESP} &
        \multicolumn{2}{c}{EST}
        \\
        \cline{2-11} 
                        & cs-cs  & cs-en & de-de & de-en & da-da & da-en & es-es & es-en & et-et & et-en \\ 
        \hlineB{3}
  Edit Distance        & 0.3215      & 0.0524      & 0.4392      & 0.0832      & 0.5650      & 0.1596      & 0.3297      & 0.0545      & 0.4121      & 0.1146      \\ 
  Word TF-IDF          & 0.2410      & 0.0023      & 0.4763      & 0.0388      & 0.5187      & 0.0398      & 0.2411      & 0.0127      & 0.3675      & 0.0097      \\ 
  Word TF-IDF (lemmas) &  -          &  -          & 0.4666      & 0.0391      & 0.5179      & 0.0404      & 0.4318      & 0.0307      &  -          &  -          \\ 
  Char TF-IDF          & 0.4043      & 0.0843      & 0.5442      & 0.1301      & 0.5809      & 0.1576      & 0.4376      & 0.1238      & 0.4838      & 0.1095      \\ 
  Char TF-IDF (lemmas) &  -          &  -          & 0.5474      & 0.1278      & 0.5801      & 0.1551      & 0.4697      & 0.1347      &  -          &  -          \\ 
  BM25                 & 0.2189      & 0.0023      & 0.3377      & 0.0050      & 0.4987      & 0.0296      & 0.1916      & 0.0073      & 0.2982      & 0.0055      \\ 
  BM25 (lemmas)        &  -          &  -          & 0.4473      & 0.0198      & 0.5125      & 0.0334      & 0.4367      & 0.0275      &  -          &  -          \\ 
        \hline
  ESCOXLM-R            & 0.1835      & 0.0195      & 0.4087      & 0.1002      & 0.3631      & 0.1095      & 0.2476      & 0.0854      & 0.2995      & 0.0374      \\ 
  mUSE-CNN             & 0.2512      & 0.0914      & 0.5606      & 0.3138      & 0.5026      & 0.1680      & 0.4176      & 0.3217      & 0.3847      & 0.0811      \\ 
  Paraph-mMPNet        & 0.2418      & 0.0464      & 0.4691      & 0.0916      & 0.4602      & 0.1148      & 0.3417      & 0.0899      & 0.3505      & 0.0635      \\ 
  BGE-M3               & 0.4285      & 0.3021      & 0.6083      & 0.3344      & 0.5839      & 0.3037      & 0.4927      & 0.3084      & 0.4726      & 0.2882      \\ 
  GIST-Embedding       & 0.3383      & 0.0854      & 0.5363      & 0.1325      & 0.5608      & 0.1348      & 0.3574      & 0.1534      & 0.3996      & 0.0597      \\ 
  mE5                  & 0.4498      & 0.3406      & 0.6122      & 0.3858      & 0.5983      & 0.3325      & 0.5021      & 0.3480      & 0.4531      & 0.2757      \\ 
  E5                   & 0.5145      & 0.4148      & 0.6639      & 0.5073      & 0.6178      & 0.4053      & 0.5557      & 0.4628      & 0.4913      & 0.2465      \\ 
  OpenAI               & 0.5141      & 0.4356      & 0.6778      & 0.5518      & 0.6173      & 0.4506      & 0.5371      & 0.4859      & 0.4969      & 0.3915      \\ 
        \hlineB{3}

    \end{tabular}
    \caption{
    Results for tasks: Czechia, Germany, Denmark, Spain, and Estonia.
    }
    \label{tab:results-appendix-mrr-b}
    \end{subtable}

    \par\bigskip

    \begin{subtable}{\textwidth}
    \centering
    \begin{tabular}{lcccccccccc}

        \hlineB{3}
        \multirow{2}{*}{Model} &
        \multicolumn{2}{c}{FRA} &
        \multicolumn{2}{c}{HRV} &
        \multicolumn{2}{c}{HUN} &
        \multicolumn{2}{c}{ITA} &
        \multicolumn{2}{c}{LTU}
        \\
        \cline{2-11} 
                        & fr-fr  & fr-en & hr-hr & hr-en & hu-hu & hu-en & it-it & it-en & lt-lt & lt-en \\ 
        \hlineB{3}
  Edit Distance        & 0.7726      & 0.0964      & 0.3791      & 0.0325      & 0.4037      & 0.0362      & 0.3919      & 0.1069      & 0.1766      & 0.0530      \\ 
  Word TF-IDF          & 0.7743      & 0.0646      & 0.4565      & 0.0058      & 0.3604      & 0.0035      & 0.1886      & 0.0164      & 0.1890      & 0.0033      \\ 
  Word TF-IDF (lemmas) & 0.7824      & 0.0810      & 0.4416      & 0.0073      &  -          &  -          & 0.4452      & 0.0142      & 0.1973      & 0.0036      \\ 
  Char TF-IDF          & 0.7956      & 0.1954      & 0.4588      & 0.0995      & 0.4249      & 0.0273      & 0.4290      & 0.1560      & 0.2054      & 0.0410      \\ 
  Char TF-IDF (lemmas) & 0.7936      & 0.1890      & 0.4657      & 0.0936      &  -          &  -          & 0.4800      & 0.1760      & 0.2118      & 0.0361      \\ 
  BM25                 & 0.7514      & 0.0484      & 0.4050      & 0.0021      & 0.3247      & 0.0030      & 0.1609      & 0.0036      & 0.1874      & 0.0033      \\ 
  BM25 (lemmas)        & 0.8042      & 0.0707      & 0.4445      & 0.0075      &  -          &  -          & 0.4425      & 0.0081      & 0.1984      & 0.0037      \\ 
        \hline
  ESCOXLM-R            & 0.6603      & 0.1098      & 0.3128      & 0.0479      & 0.2952      & 0.0311      & 0.2686      & 0.1124      & 0.1381      & 0.0209      \\ 
  mUSE-CNN             & 0.7794      & 0.3681      & 0.3790      & 0.0769      & 0.3441      & 0.0324      & 0.3732      & 0.2861      & 0.1668      & 0.0521      \\ 
  Paraph-mMPNet        & 0.7660      & 0.1624      & 0.3678      & 0.0524      & 0.3198      & 0.0219      & 0.3580      & 0.0902      & 0.1747      & 0.0196      \\ 
  BGE-M3               & 0.8454      & 0.4171      & 0.4827      & 0.2473      & 0.4496      & 0.1878      & 0.4730      & 0.3271      & 0.2212      & 0.1310      \\ 
  GIST-Embedding       & 0.8047      & 0.2011      & 0.3968      & 0.0737      & 0.3901      & 0.0306      & 0.4242      & 0.1651      & 0.1876      & 0.0355      \\ 
  mE5                  & 0.8427      & 0.4464      & 0.4734      & 0.2712      & 0.4327      & 0.2155      & 0.4825      & 0.3583      & 0.2258      & 0.1206      \\ 
  E5                   & 0.8632      & 0.5760      & 0.5074      & 0.3516      & 0.4973      & 0.3372      & 0.5384      & 0.4459      & 0.2269      & 0.1121      \\ 
  OpenAI               & 0.8721      & 0.6160      & 0.4995      & 0.3795      & 0.4715      & 0.3455      & 0.5128      & 0.4573      & 0.2295      & 0.1754      \\ 
        \hlineB{3}

    \end{tabular}
    \caption{
    Results for tasks: France, Croatia, Hungary, Italy, and Lithuania.
    }
    \label{tab:results-appendix-mrr-c}
    \end{subtable}

\end{center}
\end{table*}

\begin{table*}[htbp]
\footnotesize
\begin{center}
    {
    \caption{
    Mean reciprocal rank (MRR) for every model, evaluated in the monolingual and the cross-lingual versions of the MELO tasks.
    }
    \label{tab:results-appendix-mrr-bis}
    }
    \centering

    \begin{subtable}{\textwidth}
    \centering
    \begin{tabular}{lcccccccccc}

        \hlineB{3}
        \multirow{2}{*}{Model} &
        \multicolumn{2}{c}{LVA} &
        \multicolumn{2}{c}{NLD} &
        \multicolumn{2}{c}{NOR} &
        \multicolumn{2}{c}{POL} &
        \multicolumn{2}{c}{PRT}
        \\
        \cline{2-11} 
                        & lv-lv  & lv-en & nl-nl & nl-en & no-no & no-en & pl-pl & pl-en & pt-pt & pt-en \\ 
        \hlineB{3}
  Edit Distance        & 0.3416      & 0.0900      & 0.4275      & 0.0952      & 0.2571      & 0.0472      & 0.4911      & 0.0637      & 0.5103      & 0.1119      \\ 
  Word TF-IDF          & 0.3802      & 0.0066      & 0.4714      & 0.0460      & 0.0453      & 0.0008      & 0.5630      & 0.0143      & 0.6051      & 0.0272      \\ 
  Word TF-IDF (lemmas) &  -          &  -          & 0.4674      & 0.0435      & 0.1292      & 0.0009      & 0.5588      & 0.0216      & 0.5947      & 0.0266      \\ 
  Char TF-IDF          & 0.3845      & 0.0774      & 0.4862      & 0.1281      & 0.2876      & 0.0582      & 0.5596      & 0.1109      & 0.5855      & 0.1896      \\ 
  Char TF-IDF (lemmas) &  -          &  -          & 0.4811      & 0.1321      & 0.3272      & 0.0472      & 0.5528      & 0.1115      & 0.5860      & 0.1904      \\ 
  BM25                 & 0.3664      & 0.0054      & 0.4433      & 0.0338      & 0.0316      & 0.0002      & 0.5535      & 0.0085      & 0.5736      & 0.0236      \\ 
  BM25 (lemmas)        &  -          &  -          & 0.4320      & 0.0393      & 0.1307      & 0.0004      & 0.5482      & 0.0181      & 0.5886      & 0.0258      \\ 
        \hline
  ESCOXLM-R            & 0.1569      & 0.0276      & 0.3184      & 0.0829      & 0.1101      & 0.0267      & 0.4063      & 0.0882      & 0.4846      & 0.1312      \\ 
  mUSE-CNN             & 0.1773      & 0.0357      & 0.4255      & 0.2666      & 0.1769      & 0.1109      & 0.5141      & 0.3258      & 0.5829      & 0.3363      \\ 
  Paraph-mMPNet        & 0.2718      & 0.0343      & 0.3831      & 0.0955      & 0.1485      & 0.0492      & 0.4582      & 0.0585      & 0.5362      & 0.0996      \\ 
  BGE-M3               & 0.3842      & 0.1922      & 0.5045      & 0.3033      & 0.2662      & 0.1984      & 0.5916      & 0.3542      & 0.5878      & 0.4124      \\ 
  GIST-Embedding       & 0.3450      & 0.0603      & 0.4487      & 0.1316      & 0.2427      & 0.0716      & 0.4859      & 0.1107      & 0.5330      & 0.2017      \\ 
  mE5                  & 0.3839      & 0.1899      & 0.5059      & 0.3246      & 0.2558      & 0.2229      & 0.5836      & 0.3793      & 0.5834      & 0.4372      \\ 
  E5                   & 0.4008      & 0.1716      & 0.5650      & 0.4133      & 0.2899      & 0.3512      & 0.6220      & 0.4844      & 0.6416      & 0.5336      \\ 
  OpenAI               & 0.4103      & 0.2418      & 0.5723      & 0.4509      & 0.2946      & 0.4358      & 0.6225      & 0.5085      & 0.6339      & 0.5413      \\ 
        \hlineB{3}

    \end{tabular}
    \caption{
    Results for tasks: Latvia, the Netherlands, Norway, Poland, and Portugal.
    }
    \label{tab:results-appendix-mrr-bis-a}
    \end{subtable}

    \par\bigskip

    \begin{subtable}{\textwidth}
    \centering
    \begin{tabular}{lcccccccc}

        \hlineB{3}
        \multirow{2}{*}{Model} &
        \multicolumn{2}{c}{ROU} &
        \multicolumn{2}{c}{SVK} &
        \multicolumn{2}{c}{SVN} &
        \multicolumn{2}{c}{SWE}
        \\
        \cline{2-9} 
                        & ro-ro  & ro-ro & sk-sk & sk-en & sl-sl & sl-en & sv-it & sv-en \\ 
        \hlineB{3}
  Edit Distance        & 0.2436      & 0.0521      & 0.3321      & 0.0725      & 0.4145      & 0.0665      & 0.3254      & 0.0845      \\ 
  Word TF-IDF          & 0.2849      & 0.0261      & 0.3695      & 0.0156      & 0.4808      & 0.0083      & 0.2997      & 0.0187      \\ 
  Word TF-IDF (lemmas) & 0.2768      & 0.0332      &  -          &  -          & 0.4821      & 0.0133      & 0.3034      & 0.0191      \\ 
  Char TF-IDF          & 0.2969      & 0.1043      & 0.3961      & 0.1123      & 0.4814      & 0.0759      & 0.3848      & 0.0905      \\ 
  Char TF-IDF (lemmas) & 0.3038      & 0.1054      &  -          &  -          & 0.4850      & 0.0757      & 0.3904      & 0.0937      \\ 
  BM25                 & 0.2687      & 0.0224      & 0.3477      & 0.0120      & 0.4645      & 0.0051      & 0.2421      & 0.0125      \\ 
  BM25 (lemmas)        & 0.2621      & 0.0300      &  -          &  -          & 0.4862      & 0.0131      & 0.3002      & 0.0170      \\ 
        \hline
  ESCOXLM-R            & 0.1458      & 0.0556      & 0.2295      & 0.0429      & 0.3179      & 0.0535      & 0.2111      & 0.0681      \\ 
  mUSE-CNN             & 0.2407      & 0.1171      & 0.3118      & 0.1040      & 0.3814      & 0.0836      & 0.2837      & 0.0948      \\ 
  Paraph-mMPNet        & 0.2649      & 0.1036      & 0.2799      & 0.0639      & 0.3593      & 0.0486      & 0.2662      & 0.0611      \\ 
  BGE-M3               & 0.3167      & 0.1946      & 0.4568      & 0.3127      & 0.4999      & 0.2848      & 0.3905      & 0.1905      \\ 
  GIST-Embedding       & 0.2852      & 0.1084      & 0.3491      & 0.1041      & 0.4223      & 0.0765      & 0.3265      & 0.0772      \\ 
  mE5                  & 0.3176      & 0.2043      & 0.4632      & 0.3308      & 0.4897      & 0.2815      & 0.4001      & 0.2017      \\ 
  E5                   & 0.3314      & 0.2308      & 0.5087      & 0.3604      & 0.5339      & 0.3912      & 0.4286      & 0.2605      \\ 
  OpenAI               & 0.3383      & 0.2572      & 0.5216      & 0.4122      & 0.5400      & 0.4092      & 0.4266      & 0.2909      \\ 
        \hlineB{3}

    \end{tabular}
    \caption{
    Results for tasks: Romania, Slovakia, Slovenia, and Sweden.
    }
    \label{tab:results-appendix-mrr-bis-b}
    \end{subtable}

\end{center}
\end{table*}

\end{document}